\documentclass[letterpaper, 10 pt, journal, twoside]{IEEEtran}
\usepackage[utf8]{inputenc}
\usepackage{float}
\usepackage{caption}
\usepackage{graphicx} 
\usepackage{amsmath}
\usepackage{amssymb}
\usepackage{amsfonts}
\usepackage{mathtools}
\usepackage{bm}
\usepackage{comment}
\usepackage{subfig}
\usepackage{epsfig}
\usepackage{algorithm}
\usepackage{algpseudocode}
\usepackage{verbatim}
\usepackage{url}
\usepackage{color,soul}

\def\BibTeX{{\rm B\kern-.05em{\sc i\kern-.025em b}\kern-.08em
    T\kern-.1667em\lower.7ex\hbox{E}\kern-.125emX}}

\begin{document}   

\title{Closed-loop shape control of deformable linear objects based on Cosserat model}

\author{Azad Artinian$^{1}$, Faïz Ben Amar$^{1}$, and Véronique Perdereau$^{1}$%
\thanks{Manuscript received: February, 22, 2024; Revised July, 19, 2024; Accepted August, 14, 2024.}
\thanks{
This work was supported by  the SOFTMANBOT project,
which received funding from the European Union’s Horizon 2020 research and innovation programme under grant agreement No 869855.} 
\thanks{$^{1}$Authors are with the Institute of Intelligent Systems and Robotics (ISIR), Sorbonne University
        {\tt\footnotesize artinian@isir.upmc.fr}}%
\thanks{Digital Object Identifier (DOI): 10.1109/LRA.2024.3451368}
}

\markboth{IEEE Robotics and Automation Letters. Preprint Version. Accepted august, 2024}
{Artinian \MakeLowercase{\textit{et al.}}: Closed-loop shape control of deformable linear
objects based on Cosserat model} 

\maketitle


\begin{abstract}

The robotic shape control of deformable linear objects has garnered increasing interest within the robotics community. Despite recent progress, the majority of shape control approaches can be classified into two main groups: open-loop control, which relies on physically realistic models to represent the object, and closed-loop control, which employs less precise models alongside visual data to compute commands.
In this work, we present a novel 3D shape control approach that includes the physically realistic Cosserat model into a closed-loop control framework, using vision feedback to rectify errors in real-time. This approach capitalizes on the advantages of both groups: the realism and precision provided by physics-based models, and the rapid computation, therefore enabling real-time correction of model errors, and robustness to elastic parameter estimation inherent in vision-based approaches.
This is achieved by computing a deformation Jacobian derived from both the Cosserat model and visual data. To demonstrate the effectiveness of the method, we conduct a series of shape control experiments where robots are tasked with deforming linear objects towards a desired shape.

\end{abstract}


\section{Introduction}

\noindent Robotic manipulation of deformable objects, and specifically the shape control of deformable objects is a growing subject in the research community. Its potential applications are as diverse as common since deformable objects are everywhere, from food manipulation \cite{jones2009three} to robotic surgery \cite{leizea2015real} and garments folding \cite{matas2018sim}. One particularly important object type is Deformable Linear Objects (DLO), such as cables and beams, which are prevalent across various manufacturing sectors and have received significant attention from the community in recent years \cite{koessler2021efficient, lagneau2020automatic}. 

Multiple methods in the literature have been used to solve shape control tasks, either for deformable robots or objects. Mechanical models such as finite elements method (FEM) \cite{ficuciello2018fem, koessler2021efficient} or Cosserat-based models \cite{campisano2021closed} are powerful tools to achieve such goals. They allow the creation of a direct relationship between the inputs of the robots and the resulting object shape. However, despite being among the most precise models available, they are highly sensitive to the estimation of elastic parameters and are computationally expensive. The stiffness matrices, which are derived from these elastic parameters, require precise estimation, which can be difficult. Additionally, due to the high computational costs, most controllers using these models are open-loop \cite{ficuciello2018fem, duenser2018interactive}. 
The authors in \cite{campisano2021closed} achieved a closed-loop approach to control a deformable linear robot (DLR), but DLR control focuses more on the displacement of the tip than on the full shape of the robot and this approach can hardly be transposed to object manipulation. 
In \cite{koessler2021efficient}, the authors proposed a closed-loop controller coupled with a reduced FEM model to control a clamped DLO.

To avoid the high computational costs and achieve closed-loop control, other approaches have focused on geometrical models or model-free methods. Usually, a local deformation Jacobian is computed to create a connection between the robot's action and the object's shape. In a series of work, the authors in \cite{navarro2016automatic, navarro2017fourier}, in \cite{zhu2021vision} and \cite{lagneau2020automatic} applied visual-servoing approaches to robotic deformable object manipulation. The local deformation Jacobian matrix is initialized and then updated at each iteration. This Jacobian matrix represents the connection between the command sent to the actuators and a low-level representation of the shape of the manipulated object.
This representation can either be the 2D contour \cite{navarro2017fourier, zhu2021vision} or the positions of a set of control points on the object \cite{alambeigi2018autonomous, hu2018three}.
The Jacobian matrix is updated based on the comparison between the visual data and the commands sent on a previous time window. It therefore depends on the anterior motions of the robot. Potential motions can therefore remain unexplored or be outdated if these degrees of freedom have not been used recently. This is especially problematic for more rigid deformable objects whose dynamics depend heavily on their current pose. 
To ensure every possible motion is taken into account when computing a solution, the authors in \cite{shetab2022lattice} introduced an ARAP model in the controller. The visual data are used to update the model which, in turn, is used to predict the next potential motions and compute the Jacobian. This approach has then been extended with an optimal controller in 
\cite{aghajanzadeh2022optimal}.
However, non-physics-based geometrical models can provide unfeasible or unrealistic shapes and can sometimes lack precision, especially for large deformations where more complex modes of deformation (Legendre polynomials) are difficult to capture with such models.

In this article, we propose a novel closed-loop approach to control the shape of DLOs. The goal is to preserve the realism of mechanical models while having none of their usual drawbacks. Our approach, based on Cosserat theory, provides an efficient and precise shape controller for DLOs, even for tasks involving complex desired shapes and deformations. 
The main advantages of our approach are its accuracy and repeatability, without requiring prior precise knowledge of the manipulated object, as well as its generality since it handles complex objects made with non-isotropic, non-homogeneous composite materials. Additionally, by integrating the Cosserat equations at every iteration, we ensure that the generated shape is feasible and physically realistic. \\
The paper is organized as follows. First, we present the Cosserat framework from which are derived the constitutive ordinary differential equations (ODE) that govern the shape of the object. Then, we propose to formulate the shape control problem as a global boundary value problem (BVP) where the manipulated object is held at both limits by the robot. We then include the Cosserat ODE in our control scheme that aims at solving an initial value problem (IVP) at each iteration to solve the BVP globally. Finally, we test our approach on our experimental setup consisting of two robotic arms and several objects with different properties. 

\section{Cosserat rod model}

\noindent The Cosserat rod model is one of the most accurate mechanical models for DLOs. It has been used extensively in the soft robotics community to model \cite{boyer2020dynamics, boyer2022statics}, control \cite{rucker2010geometrically, rucker2011statics, campisano2021closed} and estimate \cite{lilge2022continuum} the shape and the dynamics of DLRs. 
In this work, we will apply the framework introduced by the authors in \cite{antman2005problems} and developed in \cite{trivedi2008geometrically, rucker2010geometrically, rucker2011statics}. It has the advantage of formulating the problem as an initial value problem. 

As seen in chapter eight of \cite{antman2005problems}, the Cosserat constitutive equations can be derived by applying static equilibrium on a section of the DLO: 

\begin{equation}
\label{eq:v et u}
\begin{aligned}
    &\dot{p} = Rv   \qquad      & v = K_{t}^{-1}R^{T}n + \tilde{v} \\
    &\dot{R} = R\hat{u} \qquad  & u = K_{r}^{-1}R^{T}m + \tilde{u} \\
    &\dot{n} = -f \\
    &\dot{m} = -\dot{p} \times n - l
\end{aligned}
\end{equation}

The Cosserat ODE govern the state $\Gamma = [p, R, n, m]^{T}$ of the object i.e. the position $p$, orientation $R$, internal forces $n$, and moments $m$ at every point of the object. $f$ and $p$ are the external forces and moments, $K_{t}$ and $K_{r}$ the translational and rotational stiffness matrices, $v$ and $u$ the linear and angular rates of change, $\hat{}$ and $\sim$ refer respectively to the bijective mapping from $\mathbb{R}^{3}$ to $\mathfrak{so}(3)$ and the undeformed rest configuration of the object. 

We consider the reference length parameter $s \in [0, L]$ for a rod of length $L$. 
If we know the vector of initial values $\Gamma_{0} = \Gamma(s=0) = [p_{0}, R_{0}, n_{0}, m_{0}]^{T}$, we solve the IVP by spatially integrating equations (\ref{eq:v et u}) with respect to $s$, and thus obtain the full state $\Gamma$ at every point of the rod.

In other words, the full state $\Gamma$ of the object can be very easily deduced if we know the initial values $\Gamma_{0}$ at $s=0$. The forward Cosserat model is well-posed \cite{rucker2011statics}, we have 12 equations (see equation \ref{eq:v et u}) as well as 12 unknowns in $\Gamma$ at every point of the rod. Therefore, the solution obtained by solving the IVP is unique.


\section{Problem Formulation}

\noindent We use the classical problem formulation based on control and objective points. 
This formulation has been used extensively in the literature \cite{ficuciello2018fem, shetab2022lattice, koessler2021efficient}.
It has the advantage of being fast to compute and easy to set up.

We define two sets of points: 

\begin{itemize}
    \item The first set of objective points which position $p_{obj}^{i}$, with $i \in [1, n_{c}]$, represent the target shape of the object.

    \item The second set of control points, selected on the object, which positions $p_{c}^{i}$, with $i \in [1, n_{c}]$, are controlled.
\end{itemize}

\begin{figure}[htb]
\centering
\includegraphics[width=0.40\textwidth]{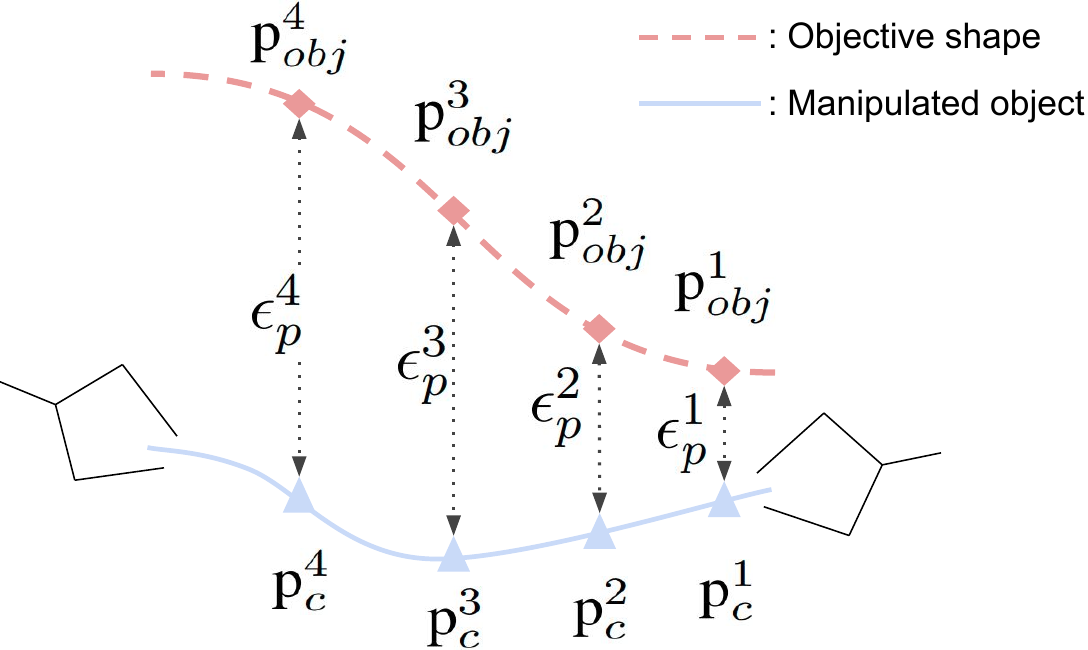}
\caption{Problem formulation: minimize $\epsilon_{p}$}
\label{fig: schema form}
\end{figure}

Each objective point is associated with a single control point and the goal of the algorithm is to minimize the position error between each objective-control point couple: $\epsilon_{p} = \begin{bmatrix} \lVert p_{obj}^{1} - p_{c}^{1} \rVert_{2} & ... & \lVert p_{obj}^{n} - p_{c}^{n}\rVert_{2}\end{bmatrix}^{T} =\begin{bmatrix} \epsilon^{1}_{p} & ... & \epsilon^{n}_{p}\end{bmatrix}^{T}$. Figure \ref{fig: schema form} illustrates a schematic representation of the problem. 

We then include the Cosserat equations that model the object to define the following BVP: find the initial values $\Gamma_{0}$ that, when integrated, provide a state of the object $\Gamma$ that minimizes $\epsilon_{p}$. 

While solving the IVP is fast and efficient and only requires a spatial integration if we know the type of boundary conditions (i.e. clamped, free at the limits, etc.), solving the BVP, i.e. finding the initial values $\Gamma_{0}$ that respect the boundary conditions is much more difficult. 
Previous approaches have relied on optimization methods to find a numerical solution \cite{rucker2011statics, boyer2022statics, azad2023optimal}, but the computational cost of these algorithms makes them difficult to include in a closed-loop control scheme.  

Our proposed approach is designed to preserve the precision of the Cosserat model while also making it less computationally expensive to include it in our closed-loop control scheme. Instead of directly solving the costly BVP, we propose to start from a known initial state that corresponds to the shape of the manipulated object at $t=0$. 
At each iteration we estimate, using sensor data, the initial values $\Gamma_{0}$ that correspond to the current shape of the object. Then, we compute a local Jacobian matrix from both the sensor data and the Cosserat equations. 
We use this Jacobian to update $\Gamma_{0}$ and solve the new IVP to obtain the updated state $\Gamma$. From this updated state, we deduce the command to send to the robots to update the shape of the actual object.
The successive computed initial values $\Gamma_{0}$ converge toward the global solution of the BVP, and therefore the shape of the manipulated object converges toward the desired shape.

\section{Closed-loop scheme}

\subsection{Deformation Jacobian based on Cosserat Model}

\noindent The deformation Jacobian matrix $J_{d}$ is a linear approximation of the Cosserat model locally. In most applications \cite{campisano2021closed, shetab2022lattice}, the authors compute the Jacobian by introducing a simulated disturbance in the input of the robots and observing the resulting simulated position of the control points. Instead, we introduce a disturbance directly in the initial configurations $\Gamma_{0}$ of the Cosserat IVP representing the object in its current shape. This allows us to update, at each iteration, $\Gamma_{0}$ and then solve the Cosserat IVP, thus ensuring the feasibility and physical realism of the resulting shape.

$J_{d}$ creates a coupling between the variations of $\Gamma_{0}$ and the variations of the positions of the control points $p_{c}^{i}$. To ensure the local approximation is correct, we compute a new Jacobian at every time step instead of updating an existing one.


Computing $J_{d}$ only requires to know the initial values $\Gamma_{0}$ corresponding to the current state of the object. 
For each component of $\Gamma_{0}$, we successively introduce a disturbance $\Delta^{j}_{d}$ with $j \in [1, m_{l}]$ and $m_{l}$ the number of components of $\Gamma_{0}$.
We then integrate the Cosserat equations from the disturbed initial conditions. 
Once we obtain the disturbed shape of the object, we compare the disturbed and current position of the control points and compute, with finite differences the column of $J_{d}$ corresponding to the disturbed component of $\Gamma_{0}$. 
By repeating this process $m_{l}$ times, we then obtain the full deformation Jacobian $J_{d}$  which provides a coupling between the variation of each dimension of $\Gamma_{0}$ and their effect on the position of the control point $p_{c}^{i}$.

The type of robotic grip on the object directly affects the Jacobian matrices and the type of boundary conditions. For a DLO grasped and controlled at both ends, we have $m_{l} = 12$, and the size of $\Gamma_{0}$ is necessarily 12, 3 for the position, 3 for the orientation, and 6 for the internal forces and moments. 

However for more realistic cases, it is often difficult to respect the clamped boundary conditions at the grasping points between the robot and the object, there can be for example friction or slippage along certain DOFs. 
Therefore, $m_{l}<12$ and the perturbations $\Delta^{j}_{d}$ must be set to zero for those uncontrollable DOFs which produces a column of zeros in $J^{i}_{d}$. 
Following the Cosserat theory, if the gripper cannot apply forces and moments along specific directions at the limits, the internal forces and moments are constant along these directions all across the rod. 

For a real-world application, the vector of initial values $\Gamma_{0}$ that corresponds to the current shape of the object is unknown. Therefore we have to estimate it at each iteration from the sensor data. In the following, we call this estimation $\hat{\Gamma_{0}}$.


\subsection{Control law}

\noindent Once $J_{d}$ is obtained, we use it to compute the new initial values 
$\Gamma_{0,k+1}$ that will be used to update the shape of the object.
We directly use the pseudo-inverse of $J_{d}$ as well as the error $\epsilon_{p}$ at step $k$ to compute the new initial values $\Gamma_{0,k+1}$ at step $k+1$: 

\begin{equation}
    \label{eq: qdes 1}
    \Gamma_{0,k+1} = \hat{\Gamma}_{0,k} + KJ^{\dag}_{d}\epsilon_{p}
\end{equation}

where $\hat{\Gamma}_{0,k}$ is the estimation of the initial values vector, $K$ is a parameter that affects the convergence rate toward the objective as well as the precision of the algorithm and $J^{\dag}_{d}$ is the Moore-Penrose inverse of the Jacobian.
The smaller $K$ is the smaller the steps between each new $\Gamma_{0}$ are. 
The algorithm will therefore require more iterations to converge toward the global solution but will also be more precise. 
Since the Jacobian is a local approximation of the Cosserat model, it is therefore more accurate when computing solutions close to the $\Gamma_{0}$ where $J_{d}$ was computed. 

Once the desired initial conditions $\Gamma_{0,k+1}$ are computed, we solve the initial value problem to acquire $\Gamma_{k+1}$. From this full state, we then deduce the
desired robotic end-effector position and orientation, and then derive a proportional control law to control the end effector's velocity and angular velocity and drive it toward the desired position and orientation, see equation \ref{vel ee}: 

\begin{equation}\label{vel ee}
    \begin{aligned}
                v_{t} &= -K_{t}e_{t}\\
                v_{a} &= -K_{a}e_{a}
    \end{aligned}
\end{equation} 

where $v_{t}$ and $v_{a}$ are the end-effector's velocity and angular velocity, $K_{t}$ and $K_{a}$ the associated gains, and $e_{t}$ and $e_{a}$ the position and orientation errors between the current end effector's pose and the new pose deduced from $\Gamma_{k+1}$.
We use the same control law for both arms.

\begin{figure}[htb]
\centering
\includegraphics[width=0.47\textwidth]{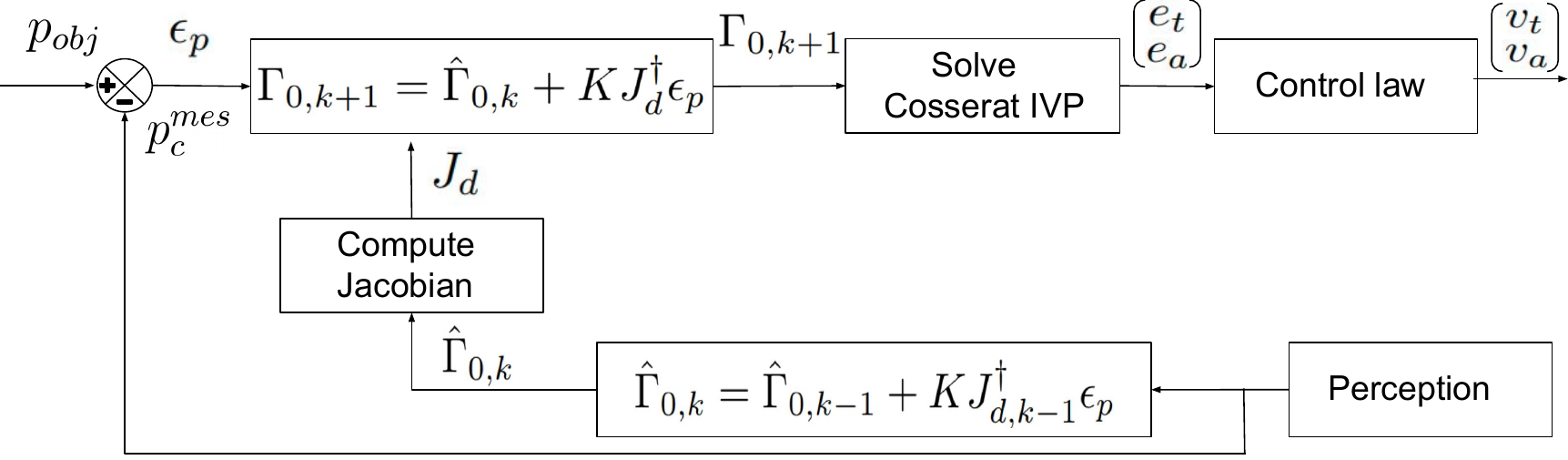}
\caption{Bloc diagram representing the control loop}
\label{fig: bloc dia2}
\end{figure}

\subsection{Perception feedback}

\noindent To compute $J_{d}$, we need to know the initial values $\Gamma_{0} = [p_{0}, R_{0}, n_{0}, m_{0}]^{T}$ that correspond to the current state of the manipulated object. It is important to have an accurate estimation of $\Gamma_{0}$ since it is the reference point from which $J_{d}$ is computed.
While the initial position $p_{0}$ and orientation $R_{0}$ are easily deduced from either the robot at the gripping point or the type of boundary conditions, the main challenge is to accurately estimate the initial internal forces $n_{0}$ and moments $m_{0}$. Two approaches could be used to estimate $\Gamma_{0}$, one based on force sensors and the other one based on vision:

\begin{itemize}
    \item \underline{Force sensor based:} The most straightforward approach. If the robot is equipped at its tip with force sensors we can simply read the force and moment values at the limits and subtract external forces and moments to obtain $n_{0}$ and $m_{0}$. The main drawback of this approach, apart from requiring two force-torque sensors, is that these measurements need to be filtered to suppress noise. This adds a delay in the loop as small noises can be amplified and generate non-negligible errors. $p_{0}$ and $R_{0}$ are then directly obtained from the robot's end-effector feedback to obtain $\Gamma_{0}$.

    \item \underline{Vision based:} This approach requires more steps but is more robust to noise or if feedback from the robot is not available. From the current error measurements of the control points $\epsilon_{p}$, we use the initial conditions and the Jacobian computed at the previous step to estimate $\Gamma_{0}$ at the current step, see equation \ref{eq: q0 1}. 

    \begin{equation}
        \label{eq: q0 1}
        \hat{\Gamma}_{0,k} = \hat{\Gamma}_{0,k-1} + KJ^{\dag}_{d,k-1}\epsilon_{p}
    \end{equation}
\end{itemize}

While the visual approach is both much easier to set up and more robust to noise, it requires precise estimation of $\Gamma_{0}$ at the first step as well as longer computation time and is also sensible to occlusions. 
We represent the control loop on the figure \ref{fig: bloc dia2} and on the algorithm \ref{alg:cl 2}.

\begin{algorithm}
\caption{Control loop at step t}\label{alg:cl 2}
\begin{algorithmic}
\State $J_{d,k-1}$ // \text{Jacobian computed at step $k-1$} 
\State $n_{0,k-1}$ // \text{internal forces computed at step $k-1$} 
\State $m_{0,k-1}$ // \text{internal moments computed at step $k-1$} \\
\\

\While{1=1}
\State $p^{mes}_{c} = getfromvision()$
\State $\epsilon_{p} = p_{obj} - p_{c}^{mes}$

\\

\State $\hat{\Gamma}_{0,k} = \hat{\Gamma}_{0,k-1} + KJ^{\dag}_{d,k-1}\epsilon_{p} $  // \textbf{estimate current initial conditions}
\\
\State $J_{d,k} = ComputeJacobian(\Gamma_{0,k}) $
\State $\Gamma_{0,k+1} = \hat{\Gamma}_{0,k} + KJ^{\dag}_{d,k}\epsilon_{p}$
\State $[e_{t}, e_{a}]^{T} = SolveCosseratIVP(\Gamma_{0,k+1})$ // \textbf{EE position and orientation error}
\State $[v_{t}, v_{a}] = ControlLaw(e_{t}, e_{a})$ // \textbf{EE velocities}
\EndWhile

\end{algorithmic}
\end{algorithm}

\subsection{Well-posed problem and stability}

The Cosserat IVP is well-posed \cite{antman2005problems}. For each vector of initial values $\Gamma_{0}$, there exists a unique solution $\Gamma$ obtained by integrating the Cosserat equations (equation \ref{eq:v et u}) from $\Gamma_{0}$. Furthermore, the solution $\Gamma$ changes continuously with $\Gamma_{0}$.
Thus, the well-posedness of the Cosserat problem guarantees a unique solution as long as the material's elastic behavior assumptions are satisfied.

Furthermore, if $\Gamma_{0}$ is updated continuously during the manipulation, then the shape of the object also varies continuously. The stability conditions can therefore be reduced to the classic visual-servoing conditions discussed in \cite{chaumette2006visual}, i.e. we can show local stability around the equilibrium points.
We define our stability conditions in a similar way as presented by the authors in \cite{gilbert2016concentric} for deformable robot control, more recently used in \cite{campisano2021closed}.
From $J_{d}$ in $s=0$, we extract the submatrix related to the variation of internal forces and moments and define our effective rigidity matrix: 

\begin{equation}
    \label{eq: rigid eff}
    M = J^{\dag}_{d}(:, 6:12)
\end{equation}

Contrary to the previous works on deformable robots, we define the rigidity matrix at $s=0$ since it is where $\Gamma_{0}$ is defined. 
We use a SVD to extract the singular values of $M$: 

\begin{equation}
    \label{eq: rigid eff}
    M = V\Sigma^{\dag}U^{T}
\end{equation}

Where $V$ and $U$ are the orthonormal matrices obtained from SVD and: 


\begin{equation}
    \Sigma^{\dag} = \mbox{diag}(\frac{\sigma_{1}}{\sigma_{1}^{2} + \epsilon_{d}^{2}}, ..., \frac{\sigma_{m_{l}}}{\sigma_{m_{l}}^{2}  + \epsilon_{d}^{2} })
\end{equation}

$\sigma_{1}, ..., \sigma_{m_{l}}$ are the singular values of $J_{d}(:, 6:12)$. They must remain positive during the manipulation to avoid instability. $\epsilon_{d}$ represents the damping term introduced in the Jacobian to avoid singular values.
This implies that the forces and moments applied by the manipulators counteract the elastic restoring forces applied by the object in $s=0$, thus ensuring the local stability around the equilibrium point where $J_{d}$ was computed. 
We therefore define the stable neighborhood $\Omega : \{ \Gamma_{0,k+1} \in \mathbb{R}^{12} \  |  \   \mbox{min}([\sigma_{1}, ..., \sigma_{m_{l}}])>0 \}$ in which each new $\Gamma_{0}$ must be computed. 

\section{Experimental Validation}

\subsection{Experimental Setup}

To validate our approach, we used two Franka Emika 7 DOF robots to manipulate various DLOs with different properties:
\begin{itemize}
    \item Object 1: A squared section DLO made of rubber with homogeneous and isotropic properties and high flexibility.
    \item Object 2: A braided steel cable with a circular section and very low flexibility. The material is neither homogeneous nor isotropic. 
    \item Object 3: A braided steel sheathed cable with a circular section and a smaller radius giving it higher flexibility. The cable is made of composite materials, not homogeneous nor isotropic either.
\end{itemize}

\begin{figure}[htb]
    \centering
    \subfloat
    {\includegraphics[width=0.3\textwidth]{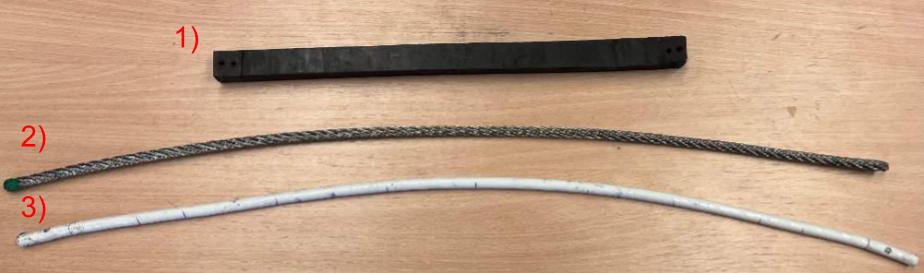}
    \label{}}
    \subfloat
    {\includegraphics[width=0.066\textwidth]{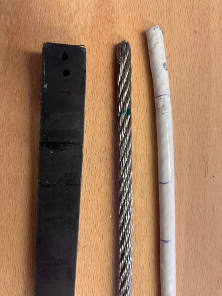}
    \label{}}
    \caption{Different objects used through the experiments}
    \label{objects}
\end{figure}

\begin{figure}[htb]
    \captionsetup[subfigure]{labelformat=empty}
    \centering
    \subfloat
    {\includegraphics[width=0.17\textwidth]{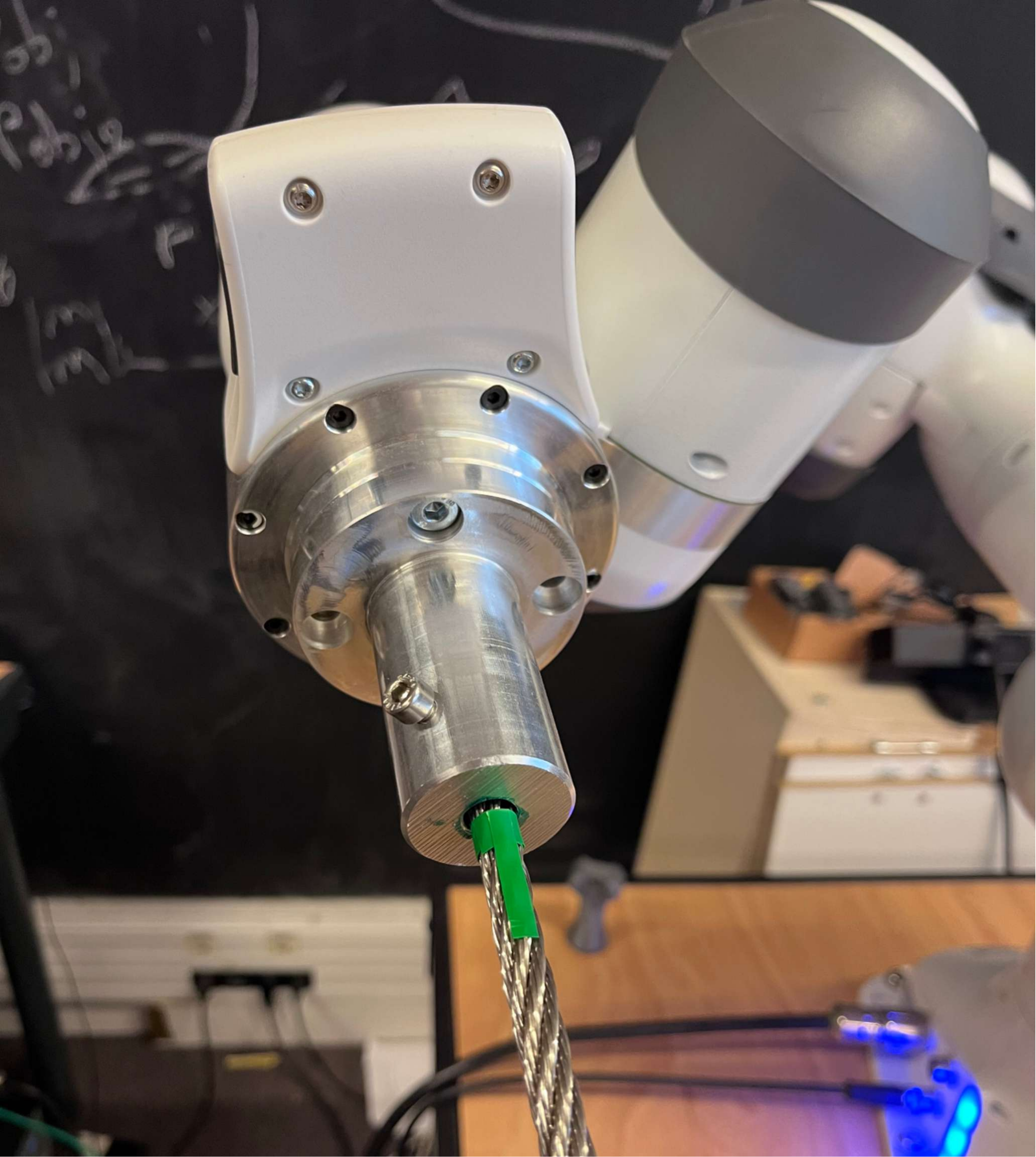}}
    \subfloat
    {\includegraphics[width=0.17\textwidth]{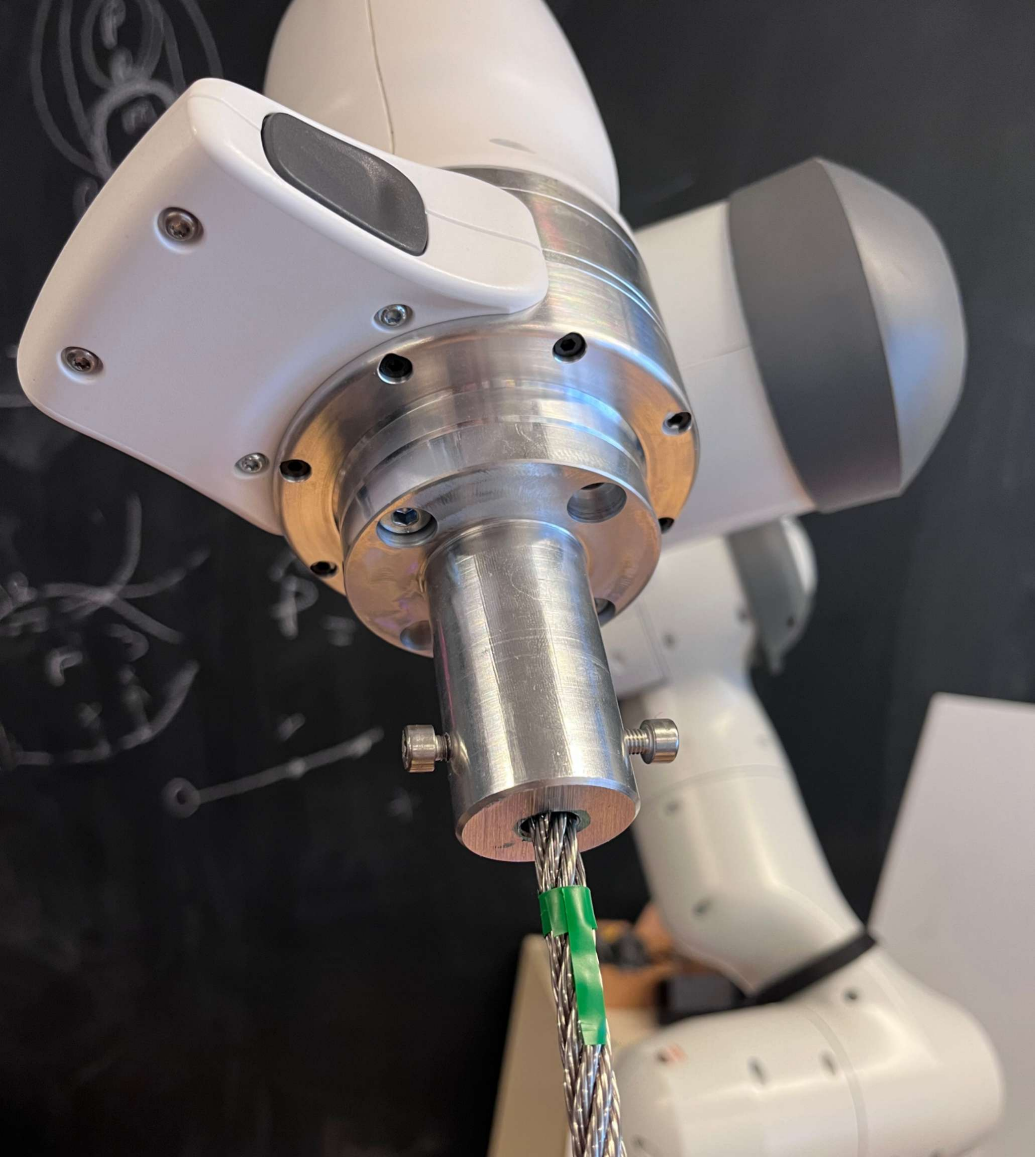}}
    \caption{Fixation of the cable before (left) and after the manipulation (right)}
    \label{fig: gliss}
\end{figure}

\begin{figure}[htb]
    \centering
    \includegraphics[width=0.40\textwidth]{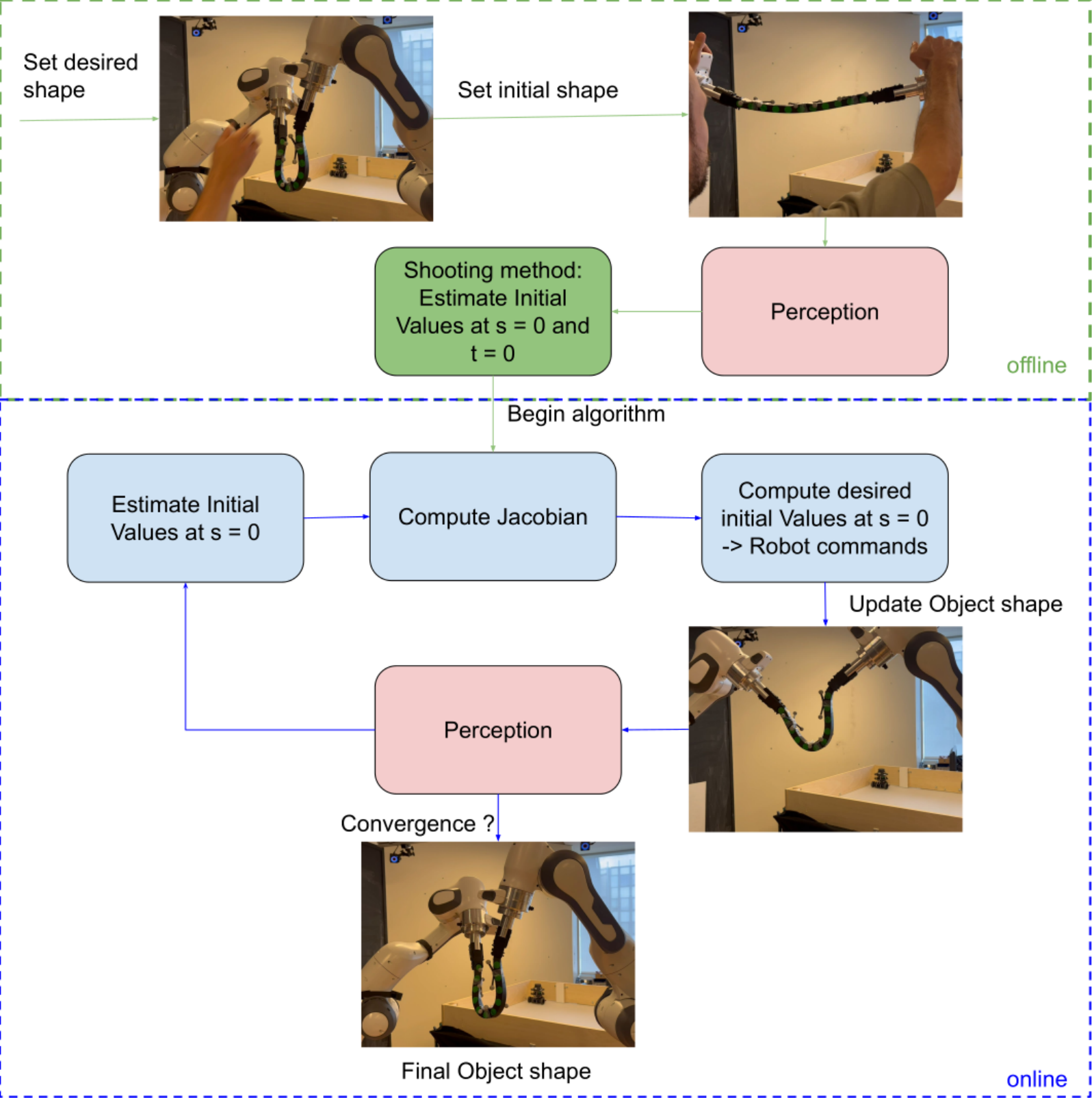}
    \caption{Experimental protocol}
    \label{fig: schema protocol}
\end{figure}

The three manipulated objects can be seen in figure \ref{objects}.
For the vision feedback, we use a set of calibrated Optitrack cameras to track markers placed on the object's control points. The approach is modulable, any vision approach can be used.
The control points must effectively capture the primary deformation modes of the object, meaning their number depends on the object's rigidity and length. For the elastic band, we use three markers, while for the longer cables, we use four. Their exact positions are not critical and vary during our tests, provided they are distributed across the entire object. We decided to use the Optitrack system both for its accuracy and robustness to occlusions. 

We use a computer with an Intel Xeon Silver 4214R CPU and the standard chipset GPU. We use ROS to communicate with the robot and the algorithm seamlessly operates at a frequency of 10 Hz, where 90 percent of the computational workload is due to the vision component. We also designed two grippers to comply with the clamped assumptions at the boundaries, i.e. no relative displacement at the contact points. 
While the conditions are respected for objects 1 and 3, there is slippage with object 2. Its effects are however compensated by the controller as the relative positions of the control points remains consistent.
The slippage during the experiments ranged between 1 to 2 centimeters along the object's axis, and up to $\pi/2$ around the axis (see figure \ref{fig: gliss}).
If the slippage were more significant, we could track the end-effector's position during manipulation and estimate the slippage amplitude by comparing the measured relative positions of the control points and the end-effector with their predicted positions. Once estimated, the slippage amplitude can be used to adjust both the boundary values of the problem and the object's length in real time.


\subsection{Protocol and initial object configuration}

\begin{figure*}[htb]
    \captionsetup[subfigure]{labelformat=empty}
    \centering
    \subfloat
    {\includegraphics[width=0.225\textwidth]{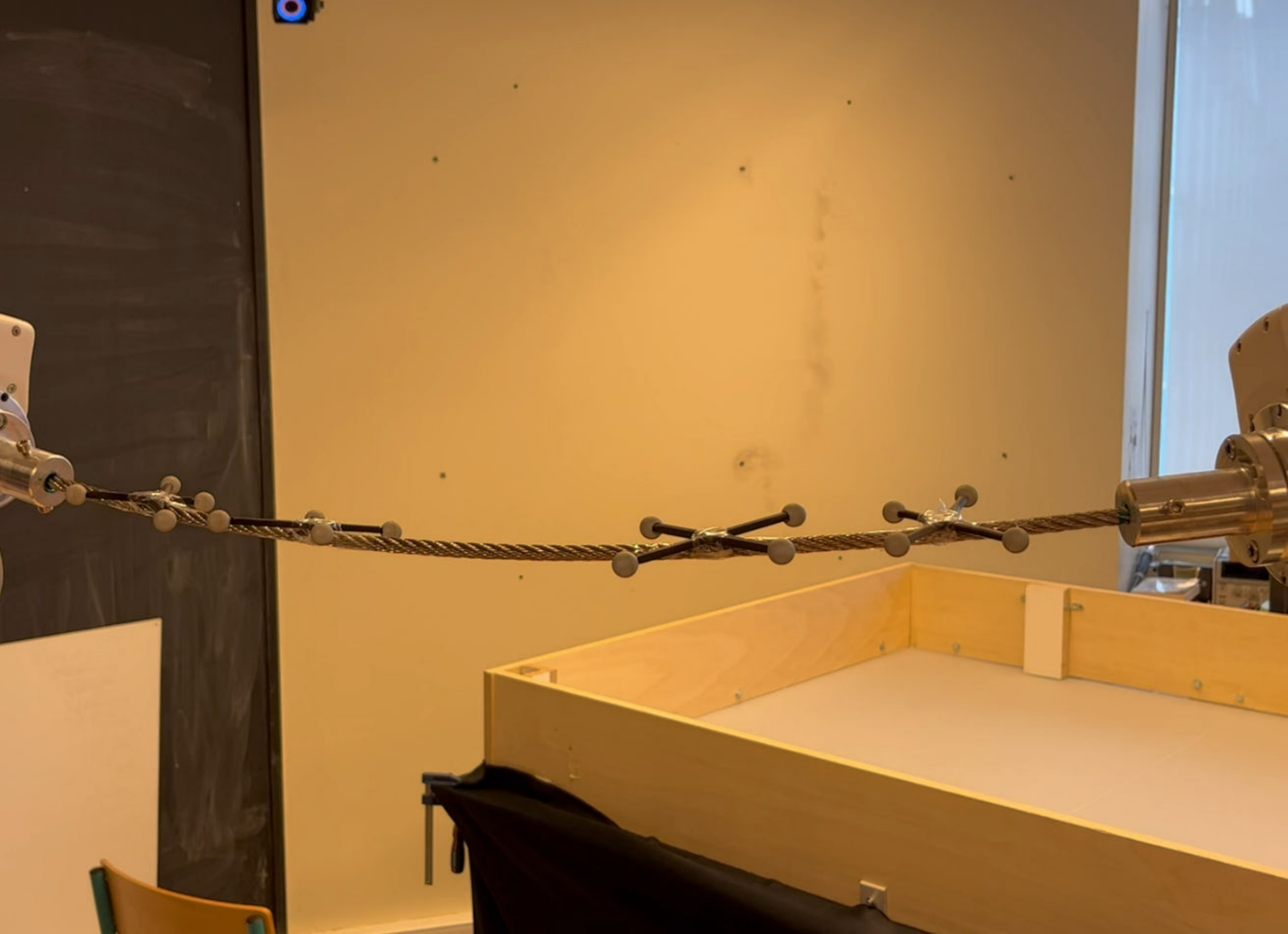}}
    \subfloat
    {\includegraphics[width=0.225\textwidth]{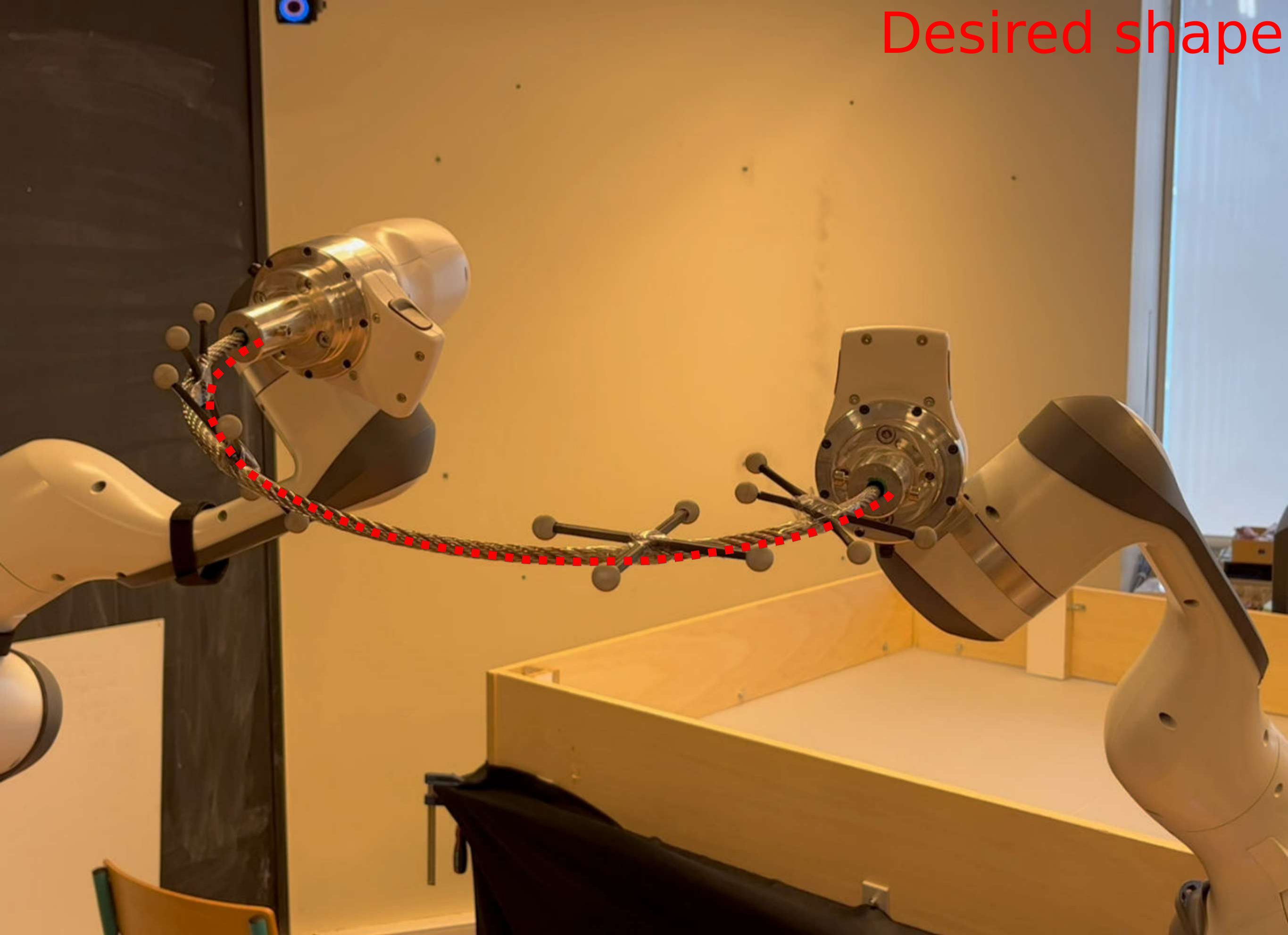}}
    \subfloat
    {\includegraphics[width=0.225\textwidth]{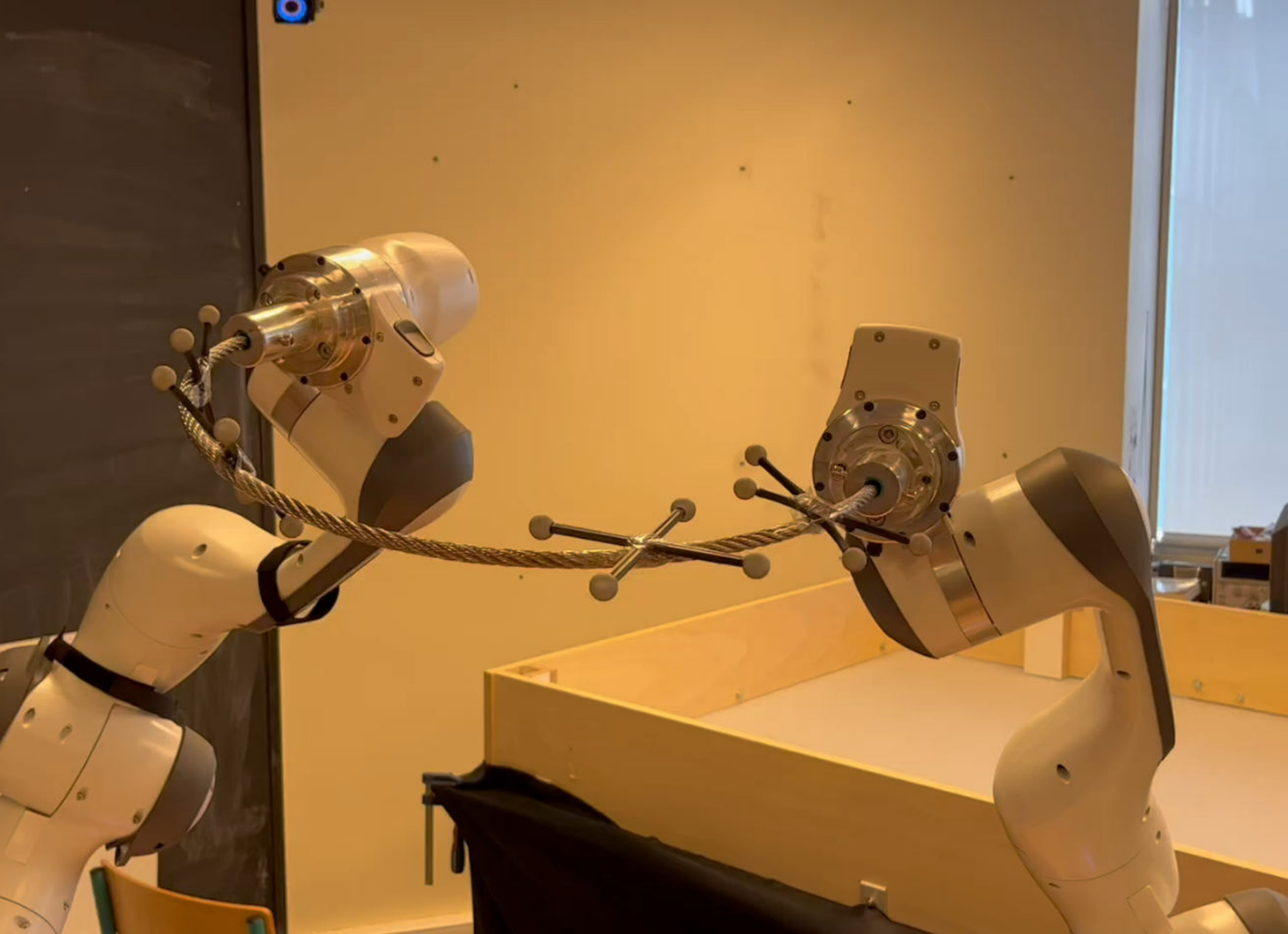}}
    \subfloat
    {\includegraphics[width=0.225\textwidth]{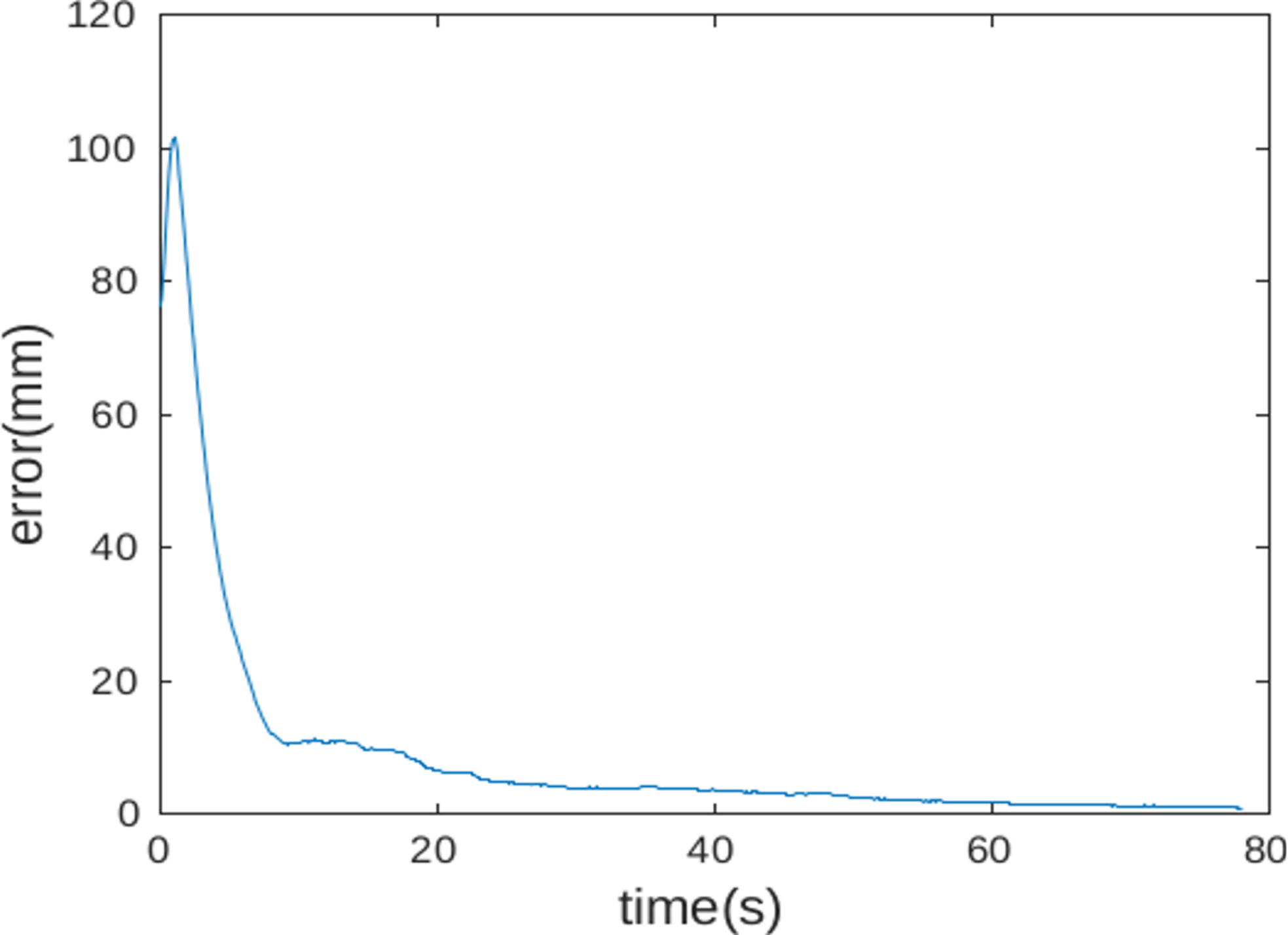}} \\[-1.9ex]

    \subfloat
    {\includegraphics[width=0.225\textwidth]{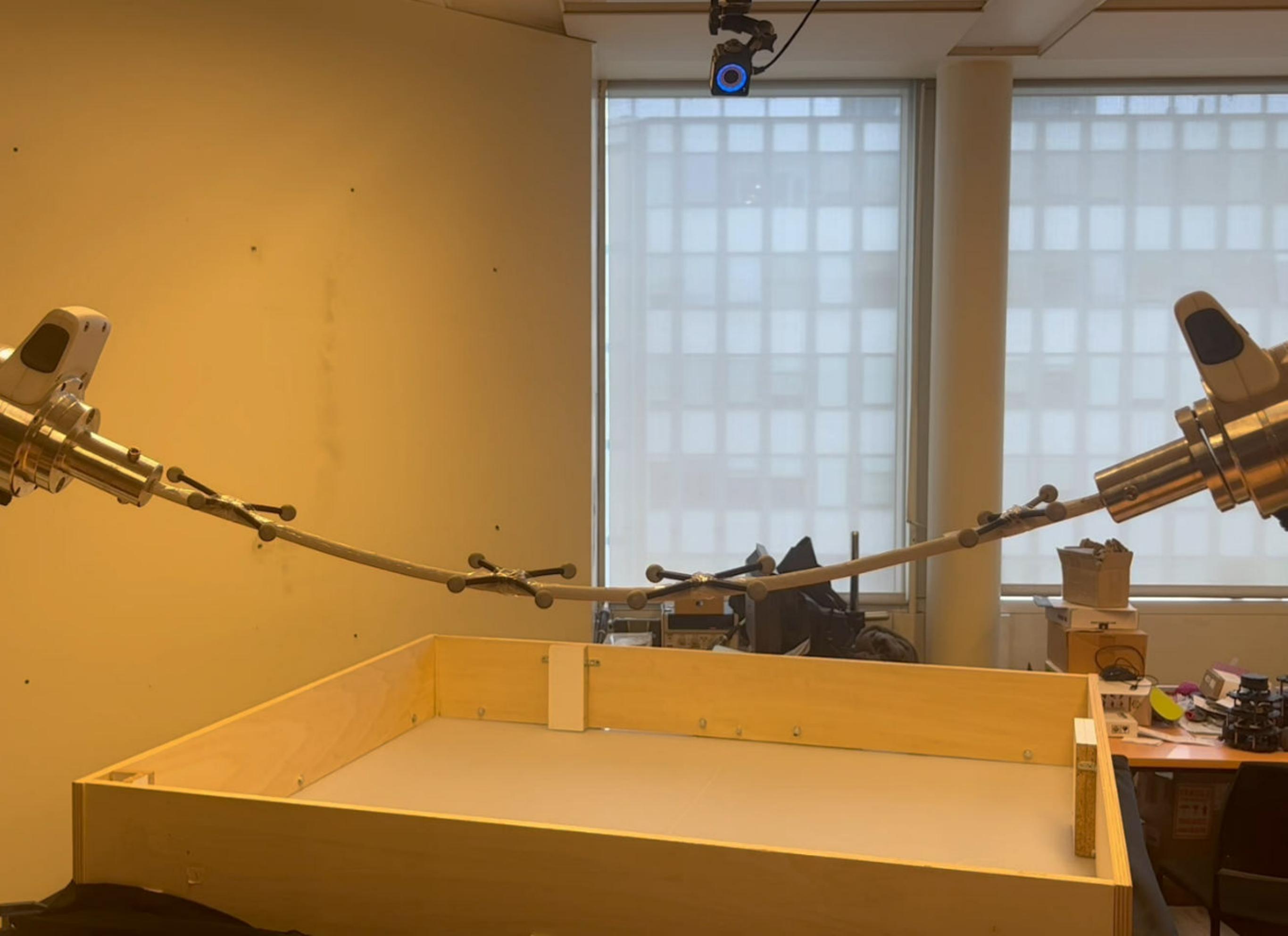}}
    \subfloat
    {\includegraphics[width=0.225\textwidth]{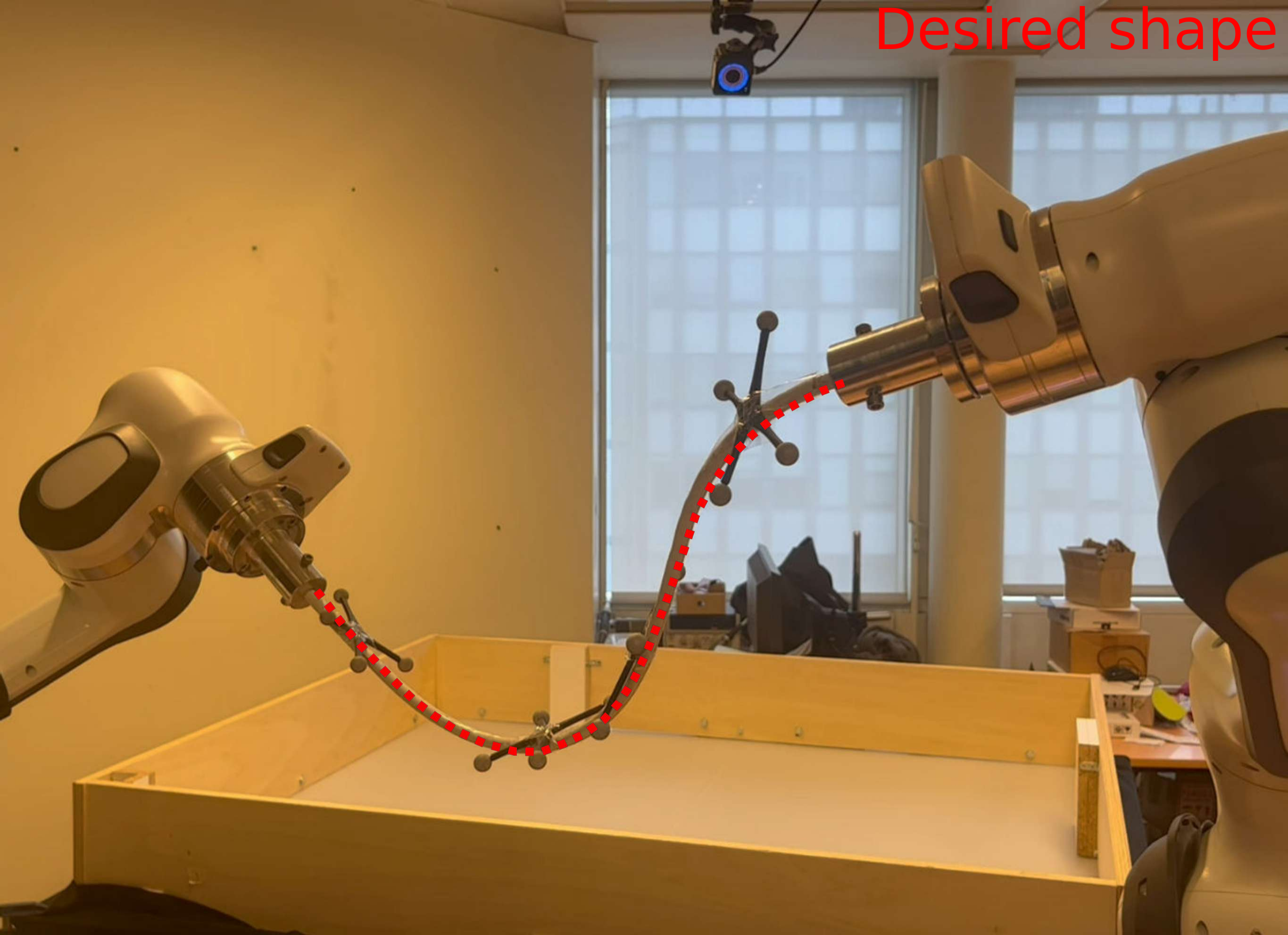}}
    \subfloat
    {\includegraphics[width=0.225\textwidth]{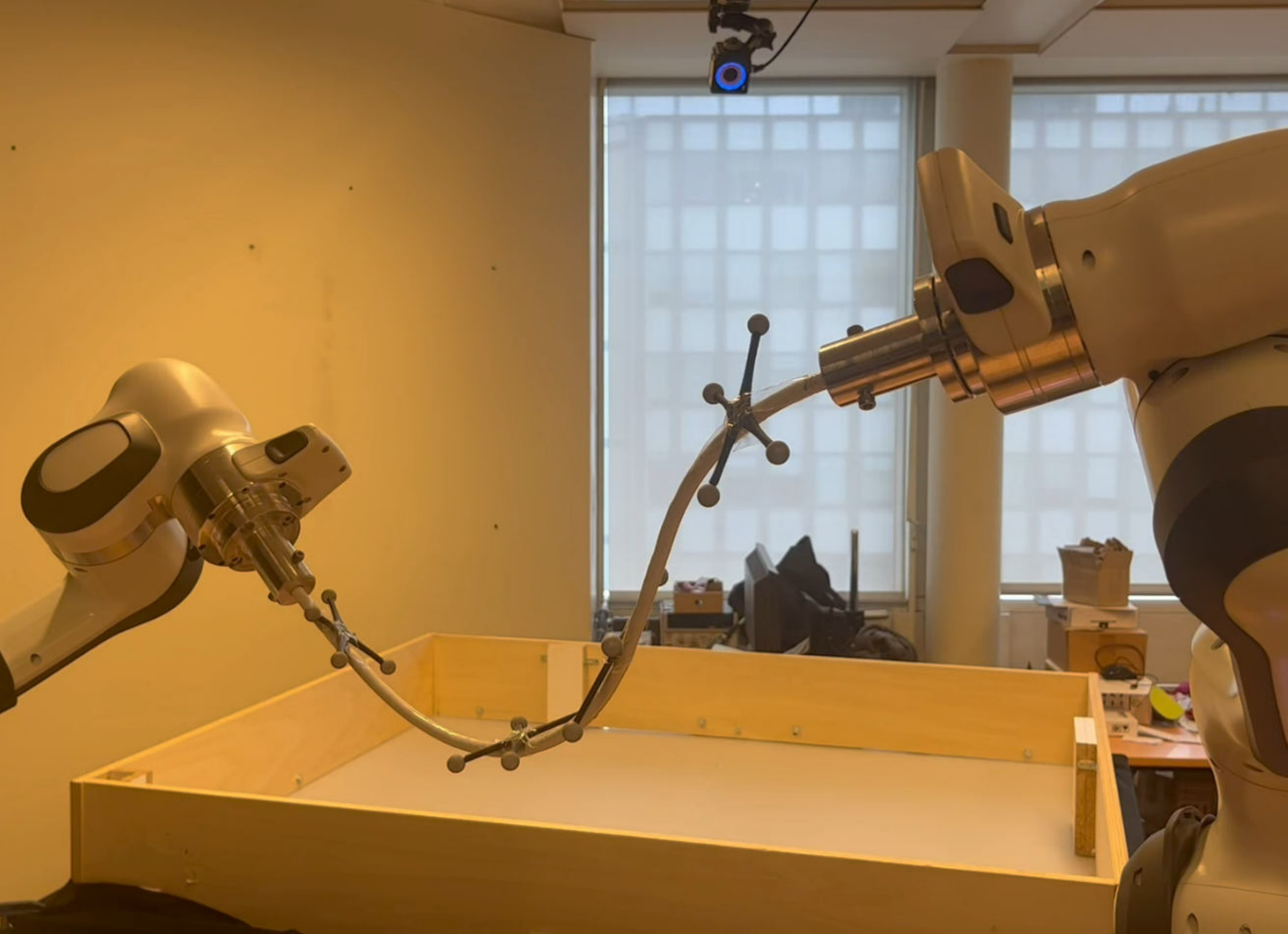}}
    \subfloat
    {\includegraphics[width=0.225\textwidth]{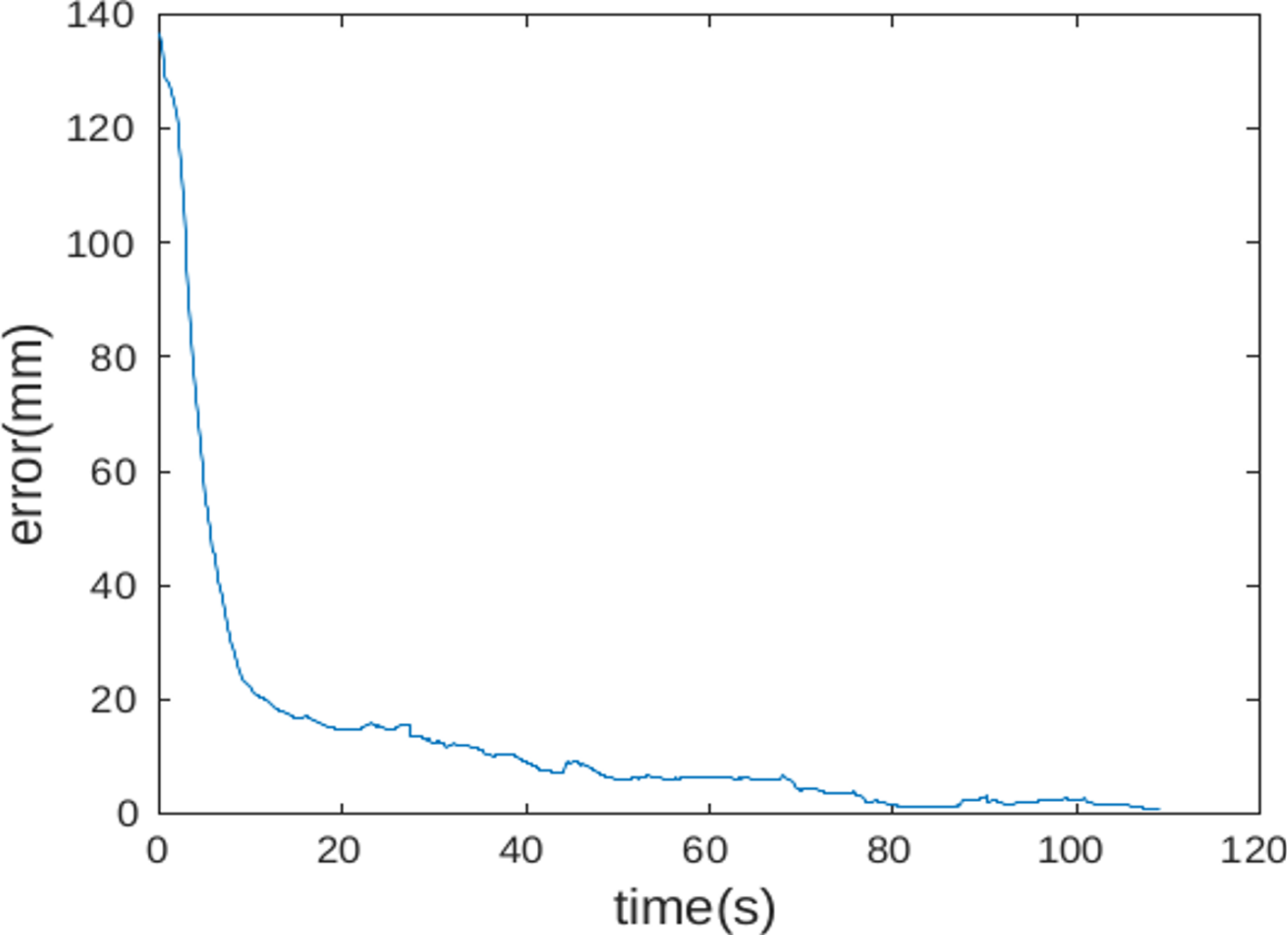}} \\[-1.9ex]

    \subfloat
    {\includegraphics[width=0.225\textwidth]{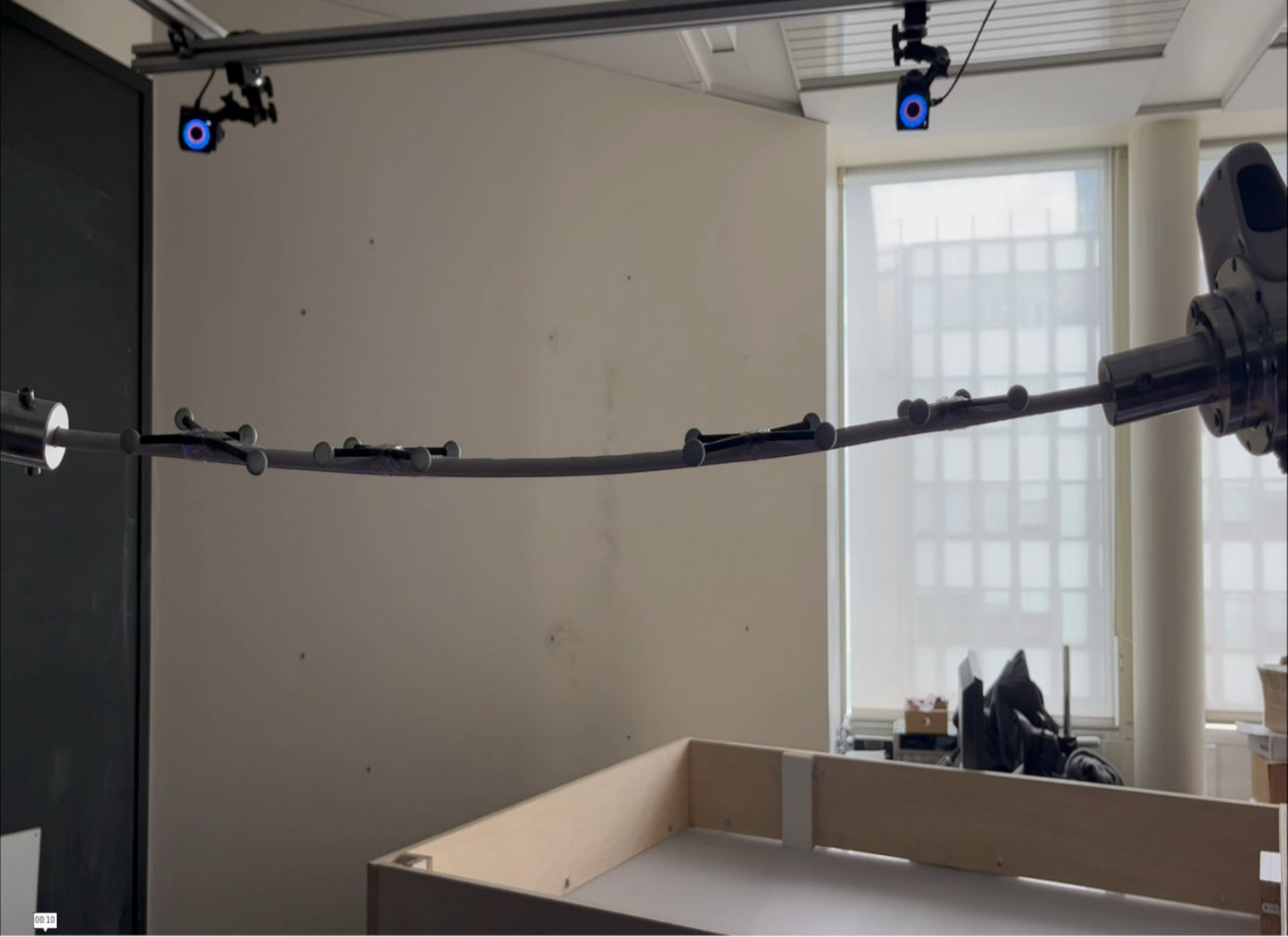}}
    \subfloat
    {\includegraphics[width=0.225\textwidth]{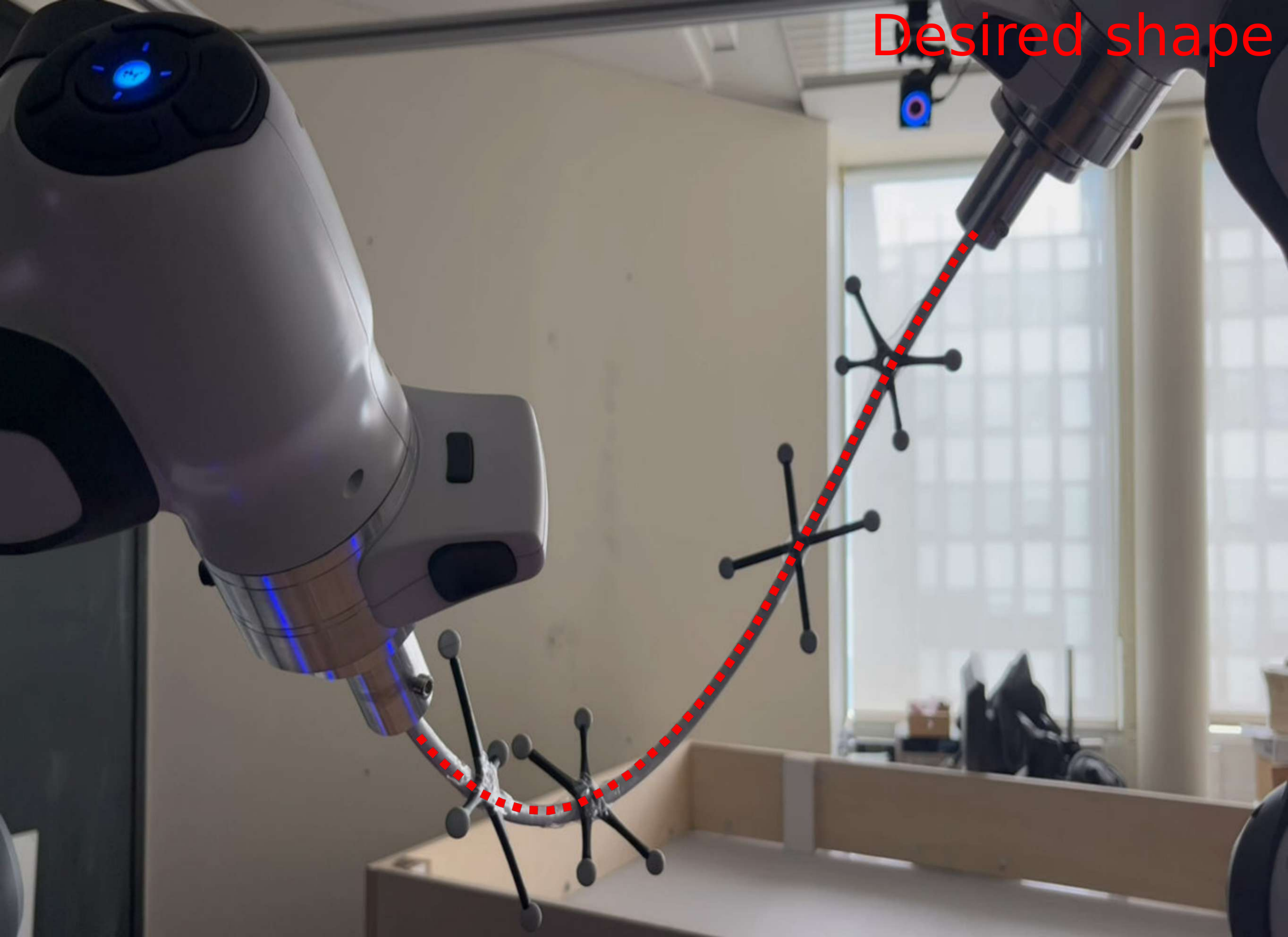}}
    \subfloat
    {\includegraphics[width=0.225\textwidth]{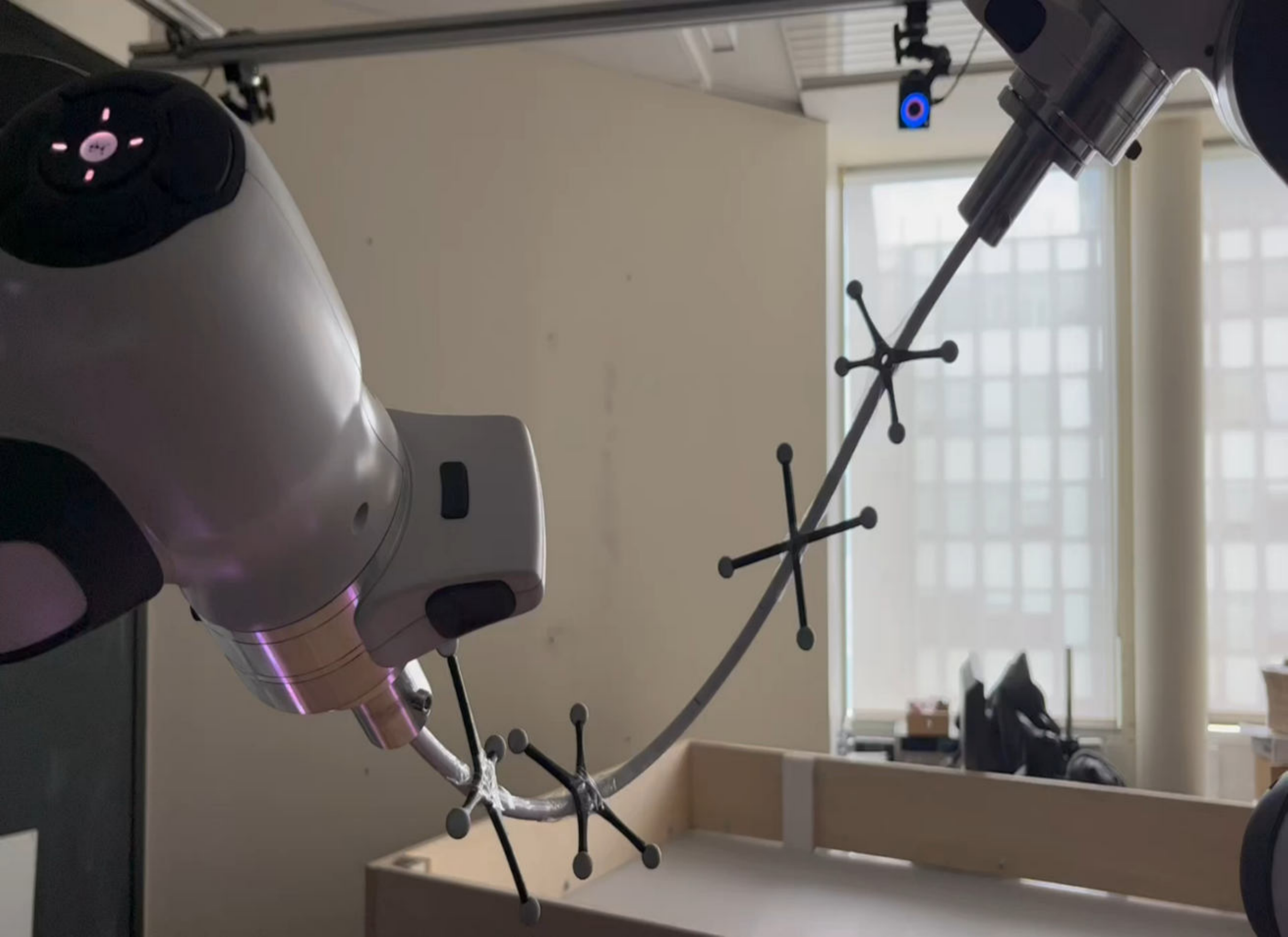}}
    \subfloat
    {\includegraphics[width=0.225\textwidth]{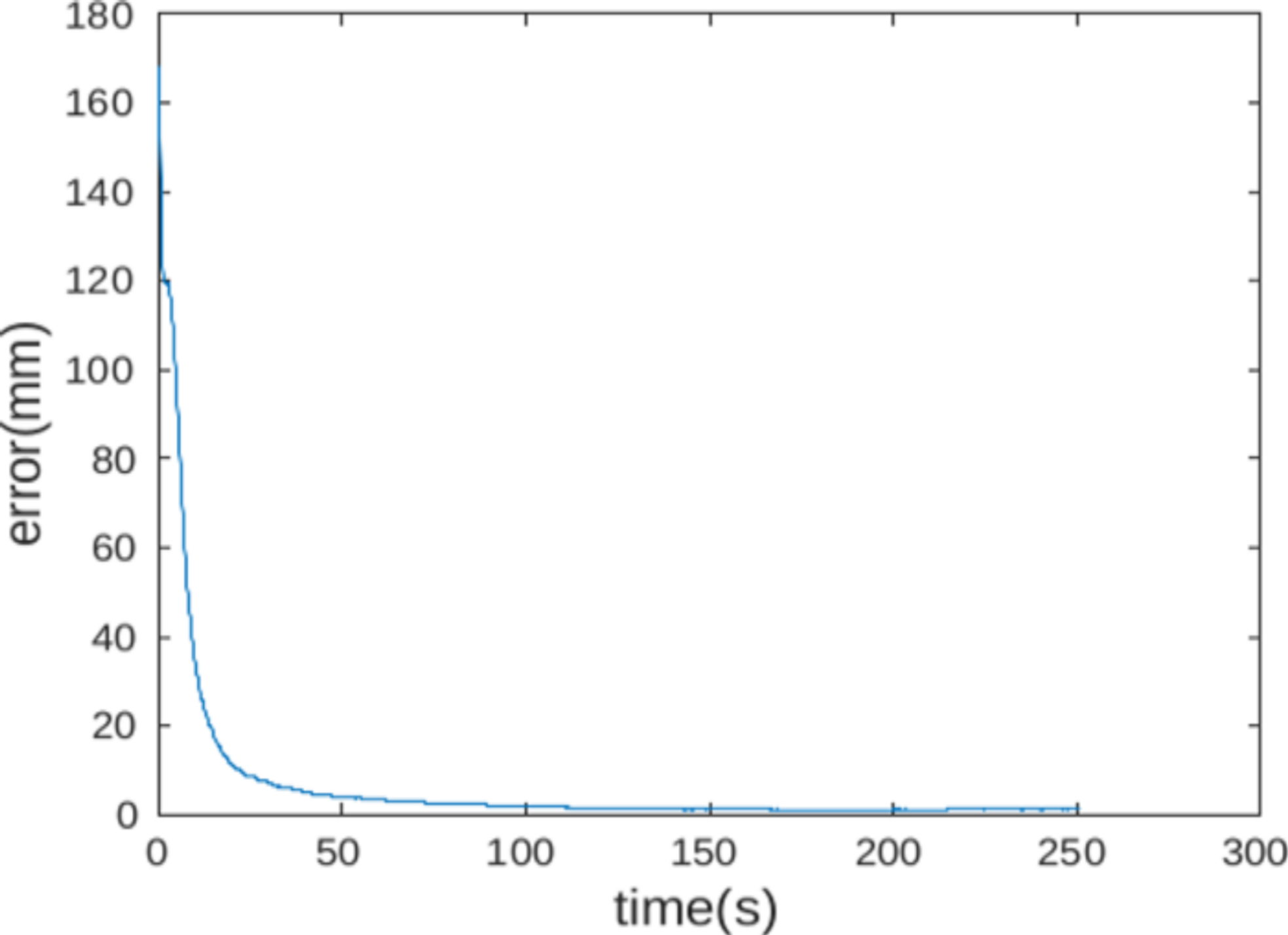}} \\[-1.9ex]

    \subfloat
    {\includegraphics[width=0.225\textwidth]{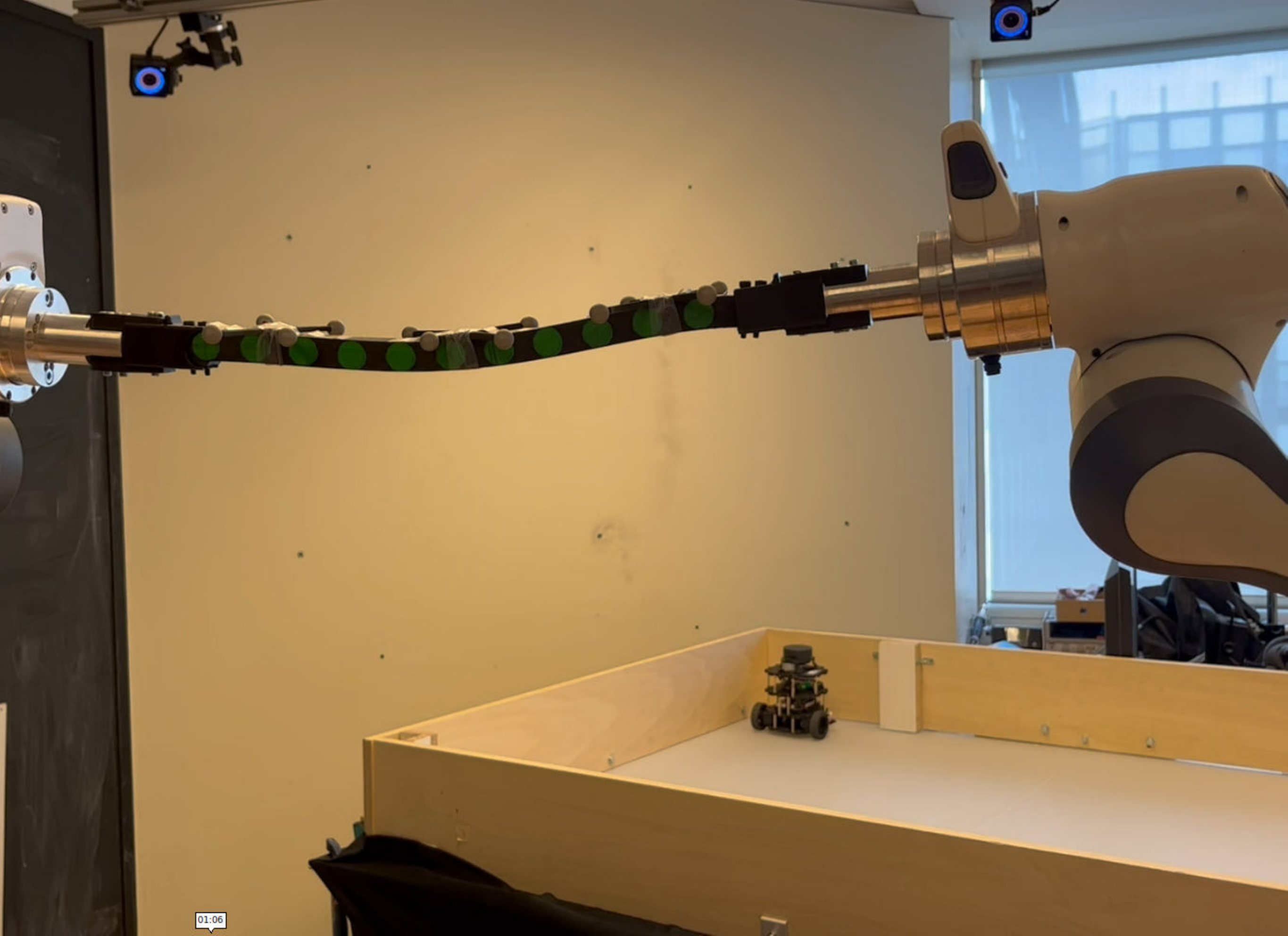}}
    \subfloat
    {\includegraphics[width=0.225\textwidth]{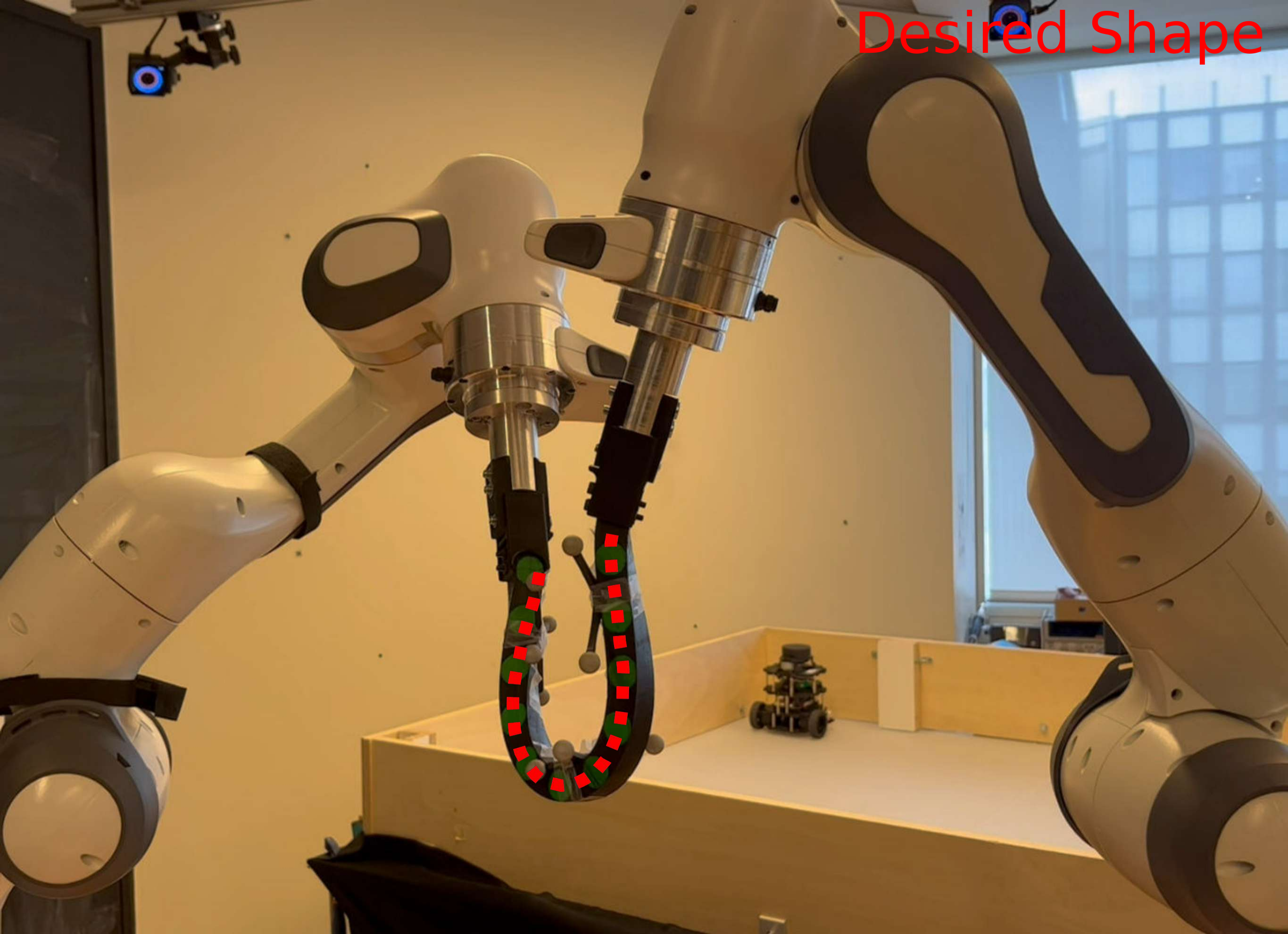}}
    \subfloat
    {\includegraphics[width=0.225\textwidth]{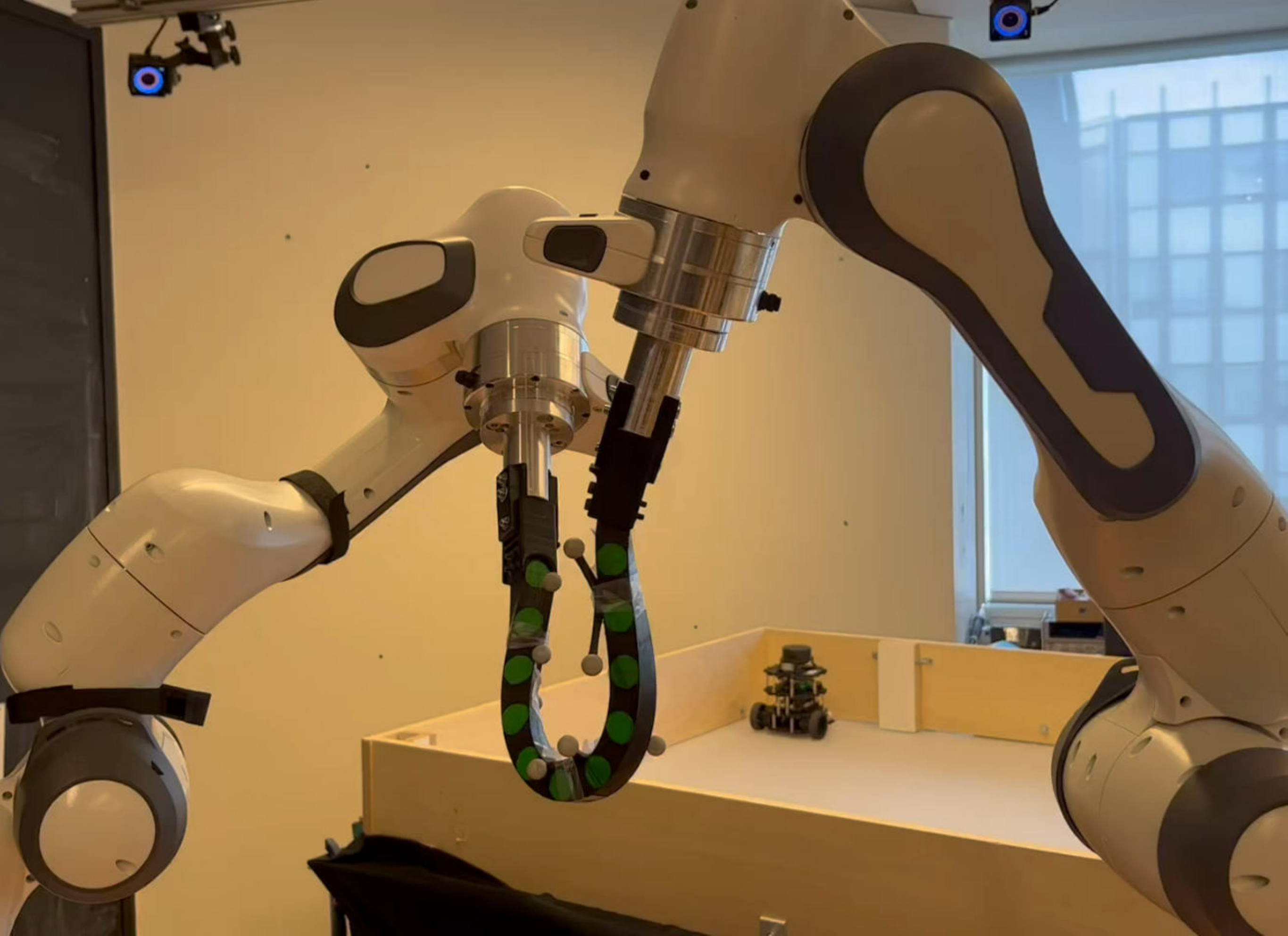}}
    \subfloat
    {\includegraphics[width=0.225\textwidth]{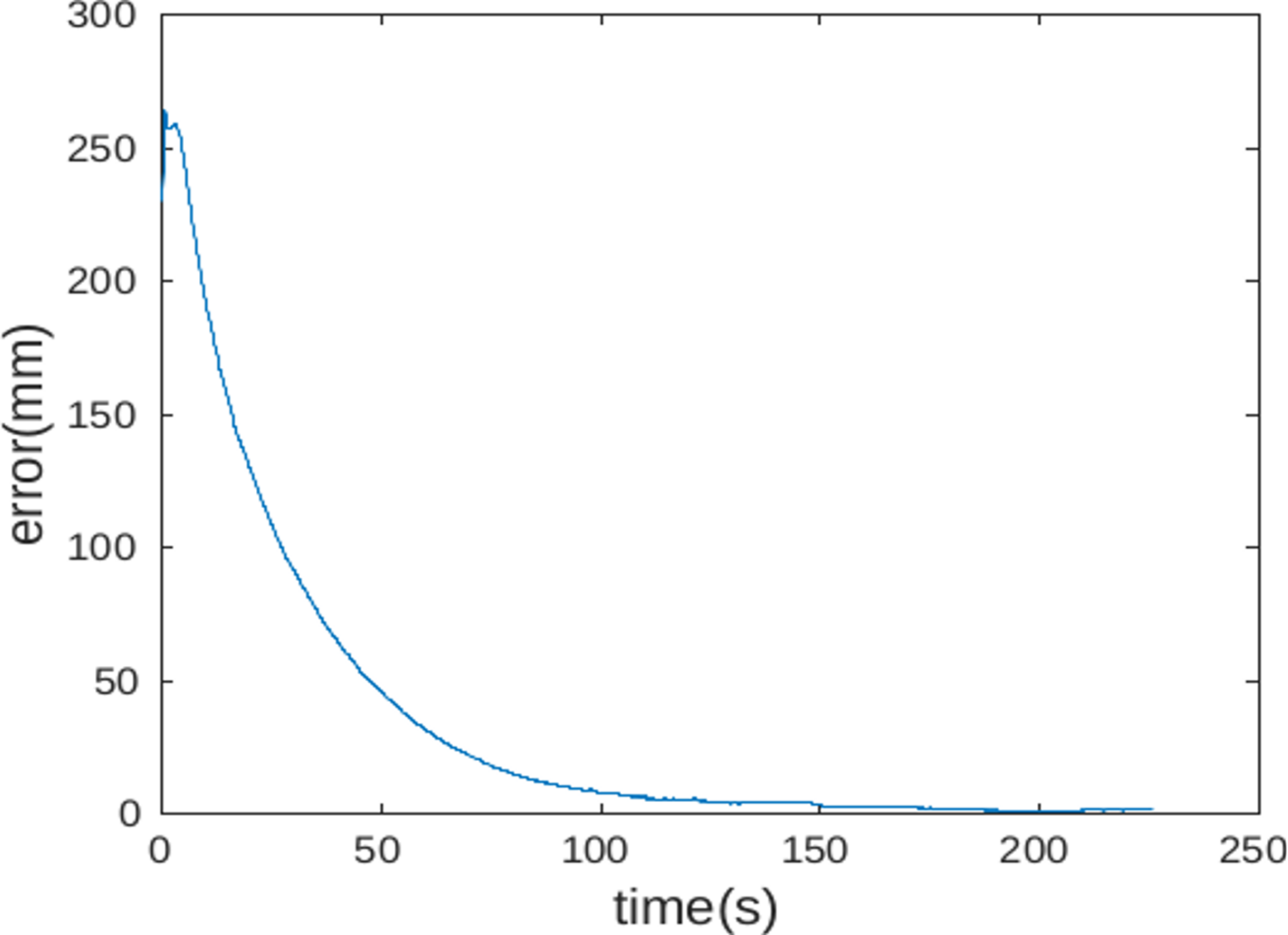}} \\[-1.9ex]

    \subfloat[Initial Shape]
    {\includegraphics[width=0.225\textwidth]{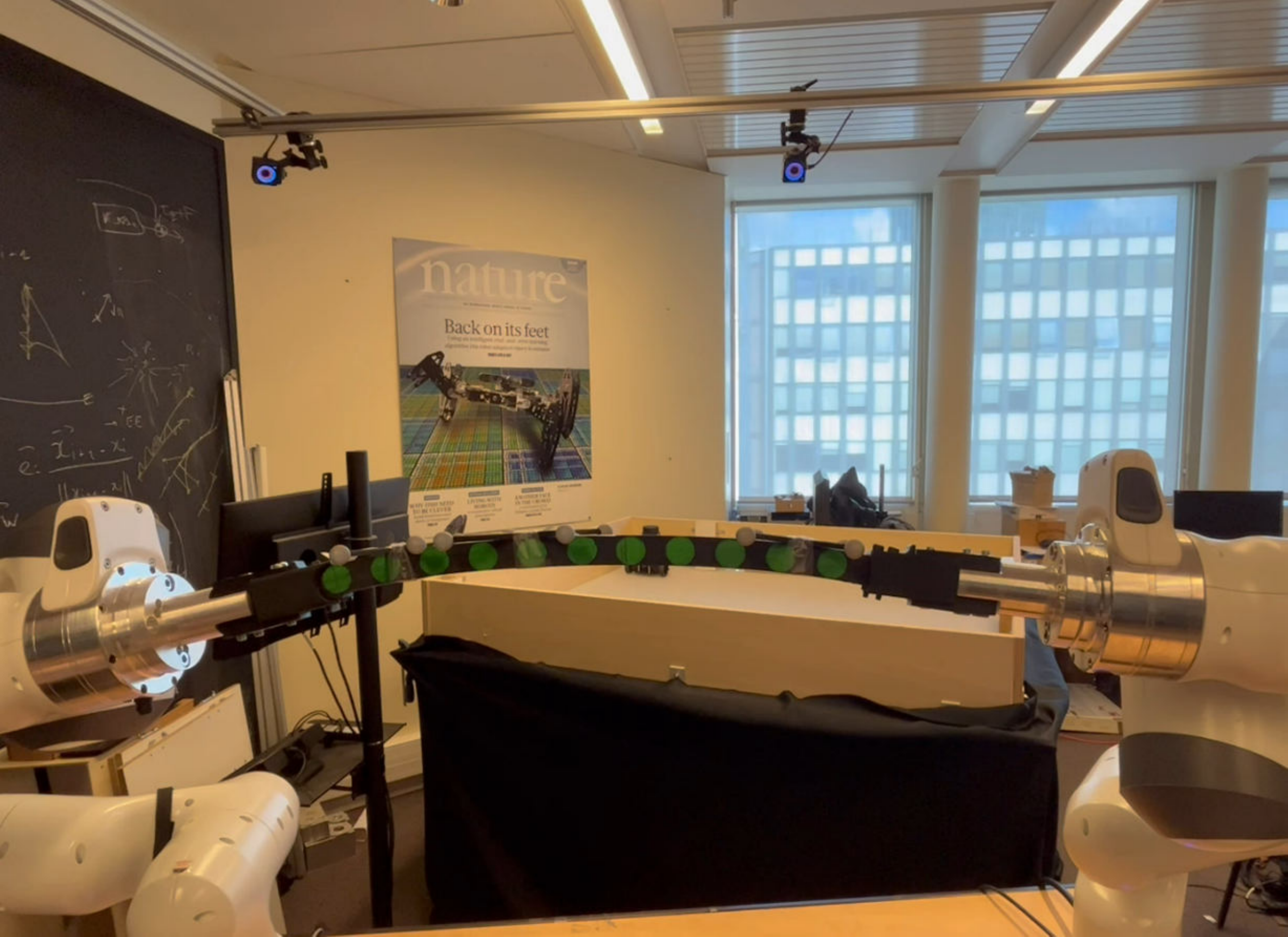}}
    \subfloat[Final Shape]
    {\includegraphics[width=0.225\textwidth]{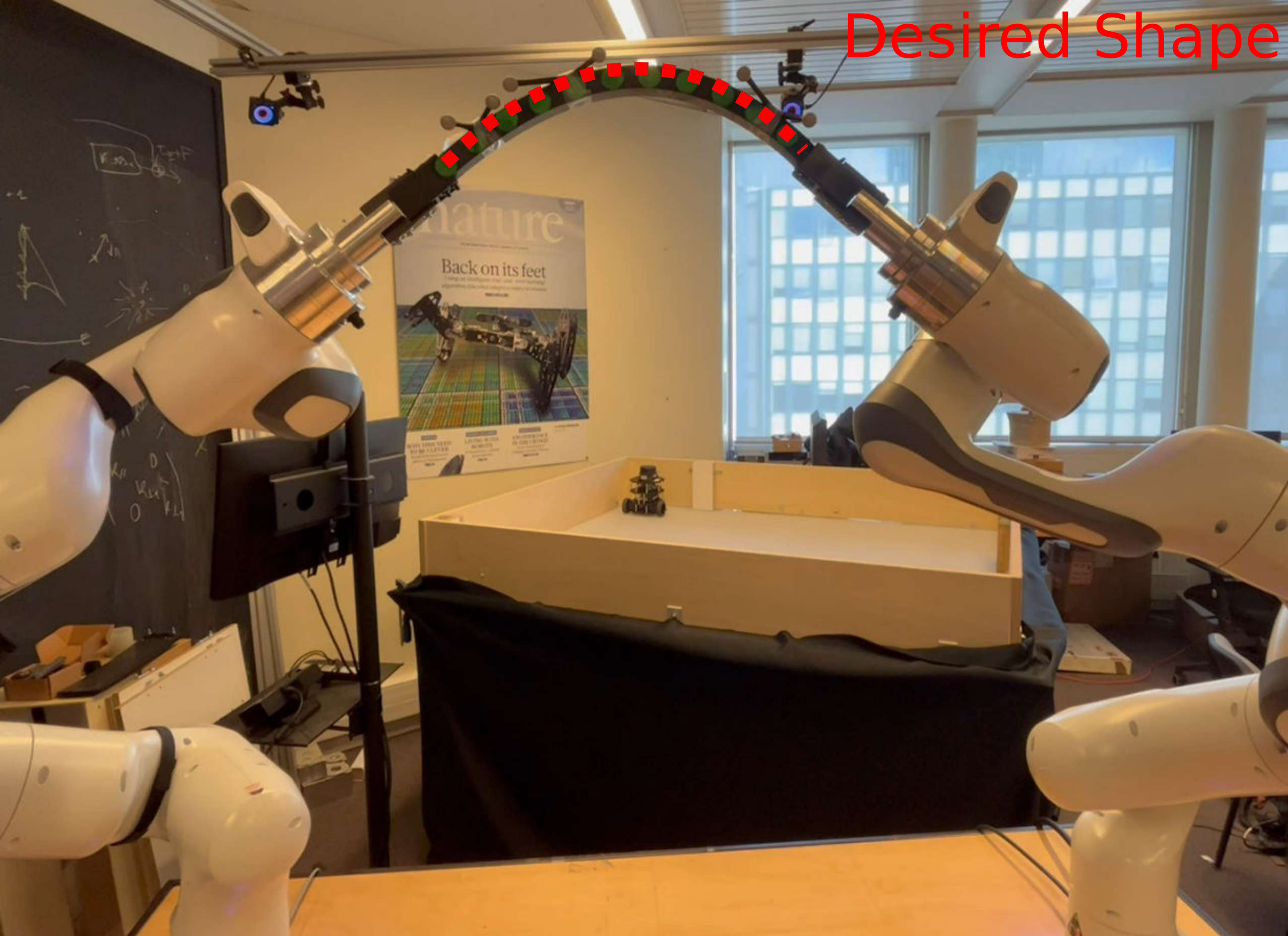}}
    \subfloat[Desired shape]
    {\includegraphics[width=0.225\textwidth]{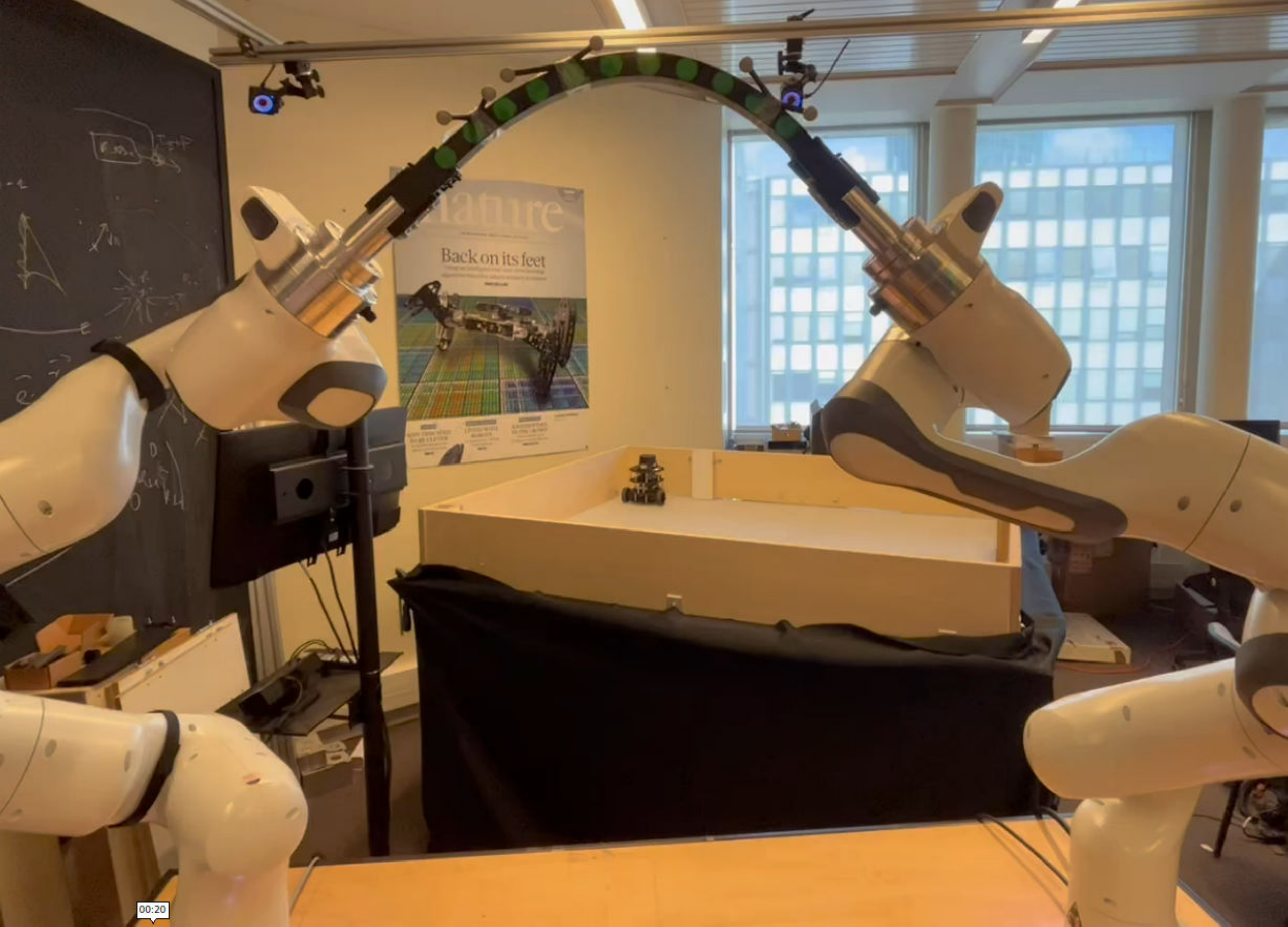}}
    \subfloat[Error over time]
    {\includegraphics[width=0.225\textwidth]{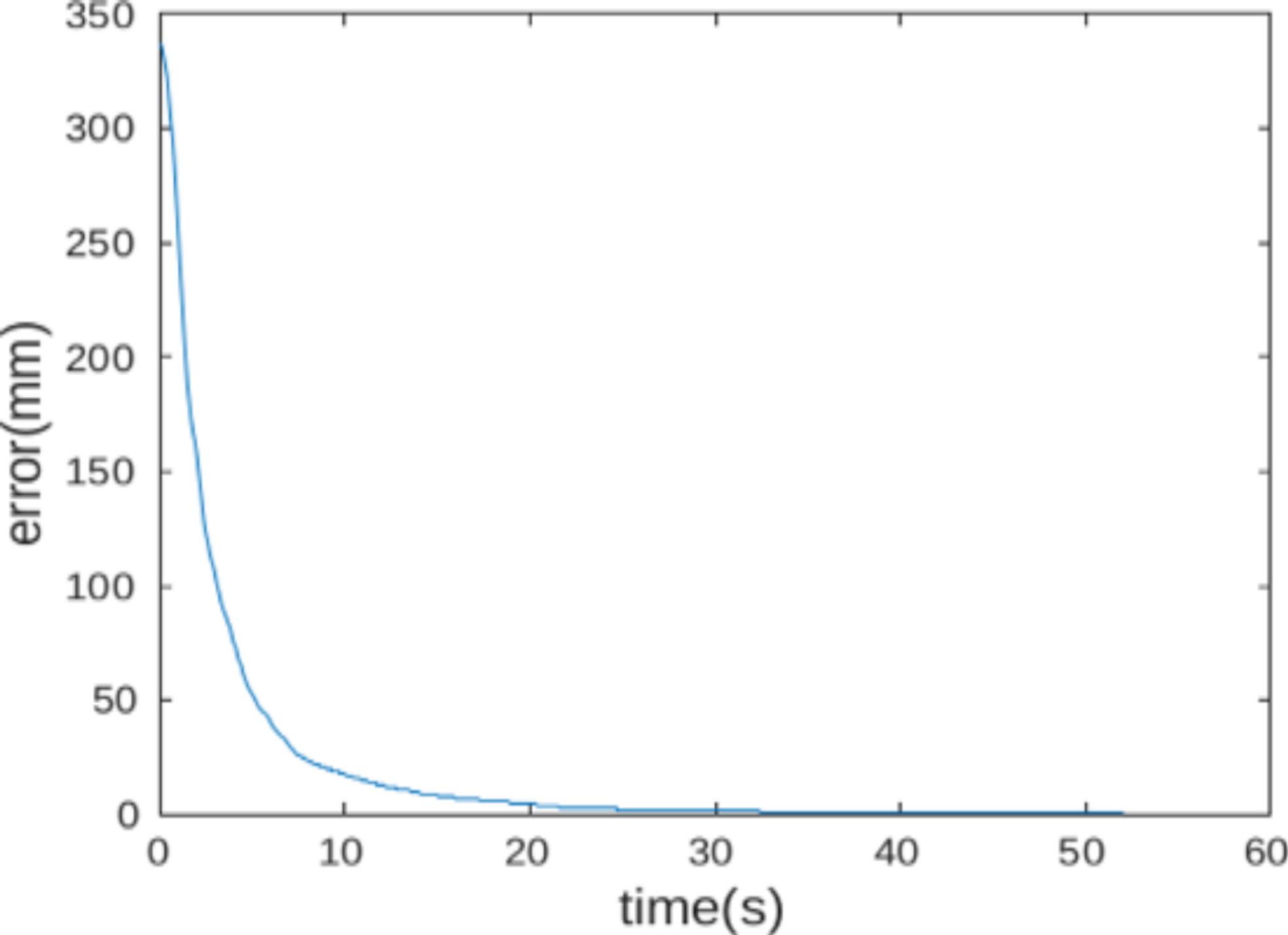}}
    
    \caption{Comparison between the initial, final and desired shapes. 
All desired shapes are set in three dimensions. 
The error over time is displayed in the left column and converges under one mm.\\ Video link: https://www.youtube.com/watch?v=UtURz1ssXO0\& }
    \label{test 1}
\end{figure*}

To evaluate the efficacy of our controller, we conducted three experiments, all involving three-dimensional shape control tasks. The first experiment focuses on bi-arm three-dimensional shape control of a DLO. We established an objective shape for the object and tasked the controller with achieving it from the initial configuration of the object. The second experiment assessed the repeatability of the approach by deforming the object towards a single predefined shape across five consecutive trials, each with a subtly varied initial object configuration.
The final experiment resembles the first one, but with altered boundary conditions. In this case, the manipulated DLO is clamped at one end and held by the robot at the other end.

We propose the following protocol, represented in figure \ref{fig: schema protocol}. First, we define the set of objective points $p^{i}_{obj}$, which represent the desired shape of the object. Next, we set the robot's end-effectors at a random initial location, activate the controller, and record the position error $\epsilon_{p}$ over time.
The initial position of the end-effectors determines the starting configuration of the manipulated object. The algorithm can begin from any initial object configuration, utilizing any method that estimates or measures the initial values $\Gamma_{0}$ at $t = 0$ corresponding to the object's initial shape.
For each test, we manually set the initial position of the end-effectors using the guiding mode of the Franka Emikas. Subsequently, we use an algorithm based on the shooting method to estimate the object's configuration. Typically, the estimation process using the shooting method takes between one and two seconds. As a result, each initial configuration varies slightly from test to test.
Almost all cases of failure come from a poor estimation of $\Gamma_{0}$ at $t = 0$, since the shooting method provides accurate estimations for the position and orientation, but is way less accurate when it comes to internal forces and moments.  

\subsection{First experiment: bi-arm shape control}

We tested our protocol on the three objects (see figure \ref{objects}) with different objective shapes. A comparison between the initial shape, final shape, and desired shape as well as the error $\epsilon_{p}$ over time is displayed in figure \ref{test 1} for a sample of all the tests. 
The error over time curves all have an exponential profile where most of the deformation happens in the first seconds. 

The elastic parameters have been estimated for object 1 (Young modulus $E = 3.2 MPa$ and Poisson ratio $\nu = 0.5$). The steel modulus for the second and third objects are $E = 180 GPa $ and Poisson ratio $\nu = 0.303$. 
However, deriving the stiffness matrices analytically for the cables proves challenging due to their mechanical characteristics. Therefore, we pre-estimate these matrices by comparing the actual deformation with predictions from the model. As a result, the obtained matrices are somewhat approximate and depend on the object's configuration.

\begin{figure}[htb]
\centering
\includegraphics[width=0.38\textwidth]{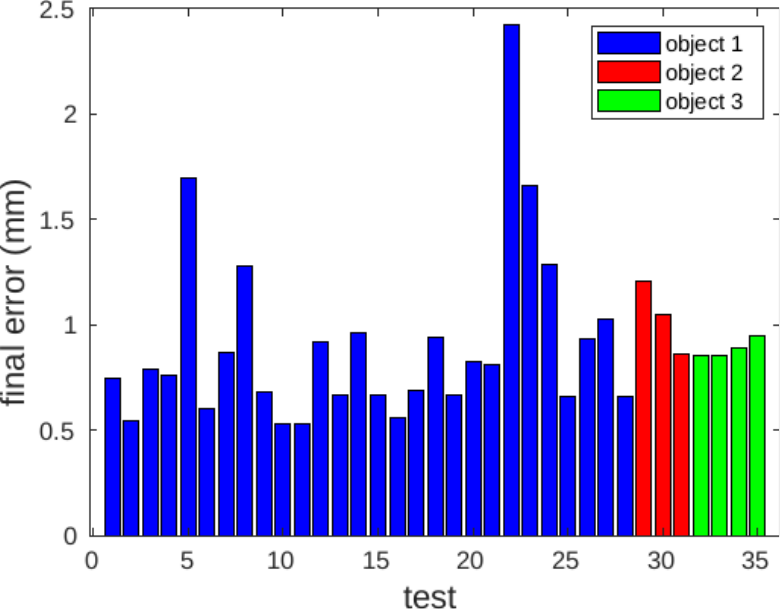}
\caption{Final error after convergence for every test}
\label{fig: total error biarm}
\end{figure}

The requirement for precise estimation of stiffness parameters is a primary limitation in physics-based models. 
Our approach effectively addresses this issue by maintaining consistent precision, regardless of the exactness of these parameters. The majority of errors stemming from inaccuracies in stiffness parameters nullify each other during Jacobian calculation using finite differences. The remaining residual error solely reflects the modeling inaccuracies introduced by the disturbance $\Delta^{j}_{d}$ which is negligible.
We conducted a total of 35 tests with different desired and initial shapes. The final errors $\epsilon_{p}$, obtained after convergence when the algorithm is stopped, are displayed in figure \ref{fig: total error biarm}. Out of our 35 tests, only 6 converge with a final error $\epsilon_{p}$ larger than 1 mm. We also had 4 cases of failure. We considered a test failed when the final error was larger than 3 mm. The causes for these cases of failure were either a poor estimation of $\Gamma_{0}$ at $t=0$ or an unattainable objective shape. 

\begin{figure}[htb]
    \centering
    \subfloat
    {\includegraphics[width=0.249\textwidth]{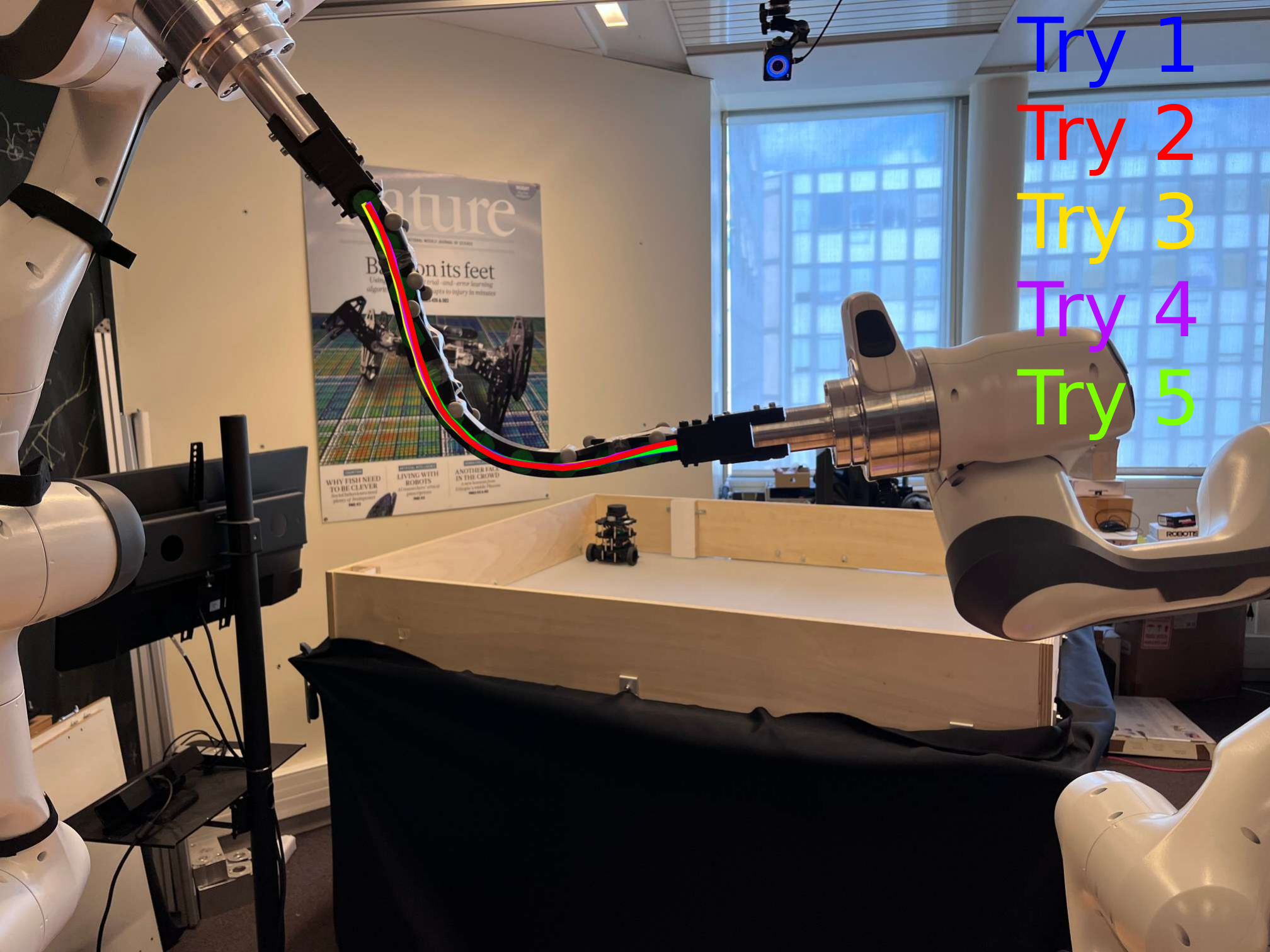}}
    \subfloat
    {\includegraphics[width=0.249\textwidth]{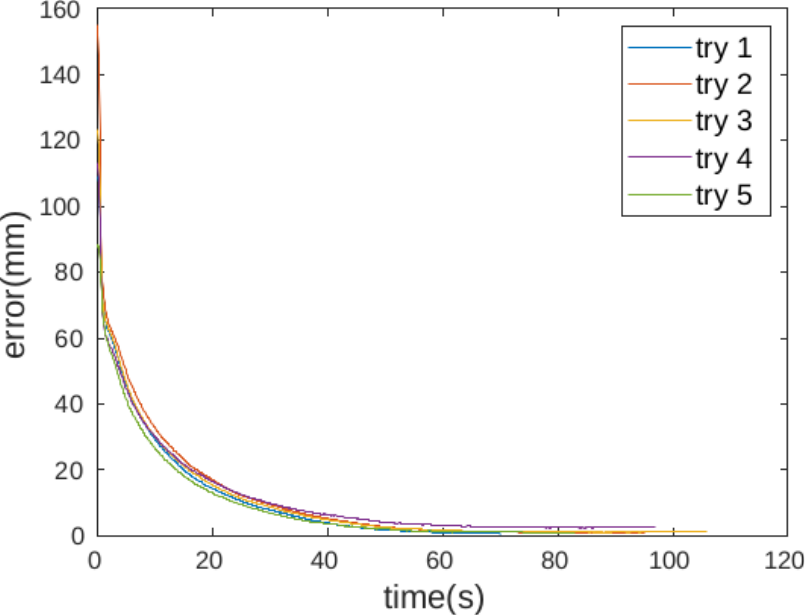}}
    \caption{Repeatability: Final shapes and error over time of five successive tests}
    \label{fig: repeat 2}
\end{figure}
\subsection{Second experiment: repeatability}
To test the repeatability of our approach, we defined a target shape and attempted to reach it in five consecutive tests.
Since the initial shape of the object is set by hand, it is slightly different every time and the error $\epsilon_{p}$ at $t=0$ varies between 90 and 160mm among the successive tests. 
The final shape, as well as the error $\epsilon_{p}$ for these five repeated tests are represented in figure \ref{fig: repeat 2}. 
The profile of the error curves and the final shapes are very similar for each test. Except for the 4th test, the object converges toward the desired shape with a sub-millimeter error at every test, regardless of the initial shape of the object.

\subsection{Third experiment: different boundary conditions}

\begin{figure*}[hbt]
    \captionsetup[subfigure]{labelformat=empty}
    \centering

    \subfloat[Initial Shape]
    {\includegraphics[width=0.22\textwidth]{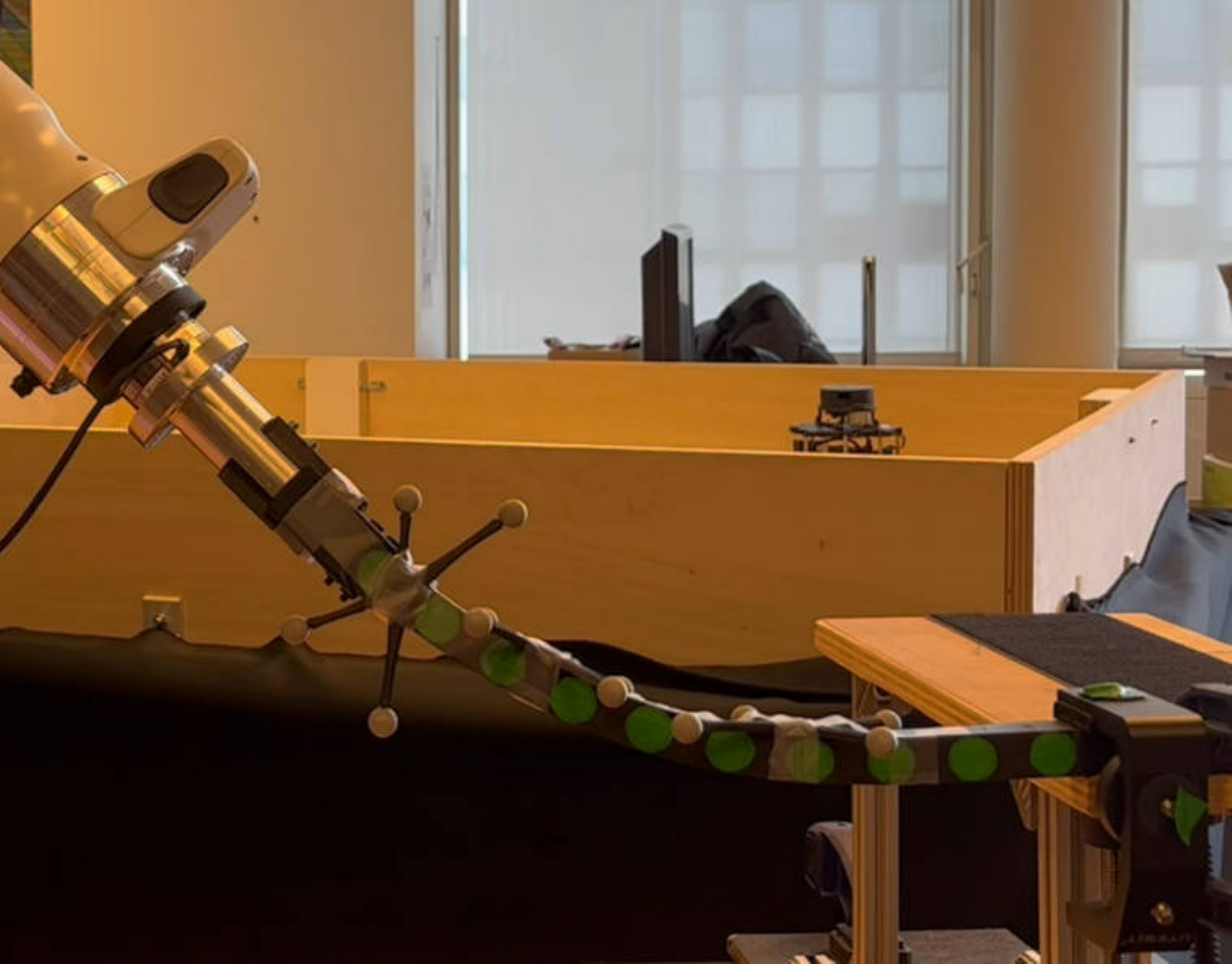}}
    \subfloat[Final Shape]
    {\includegraphics[width=0.22\textwidth]{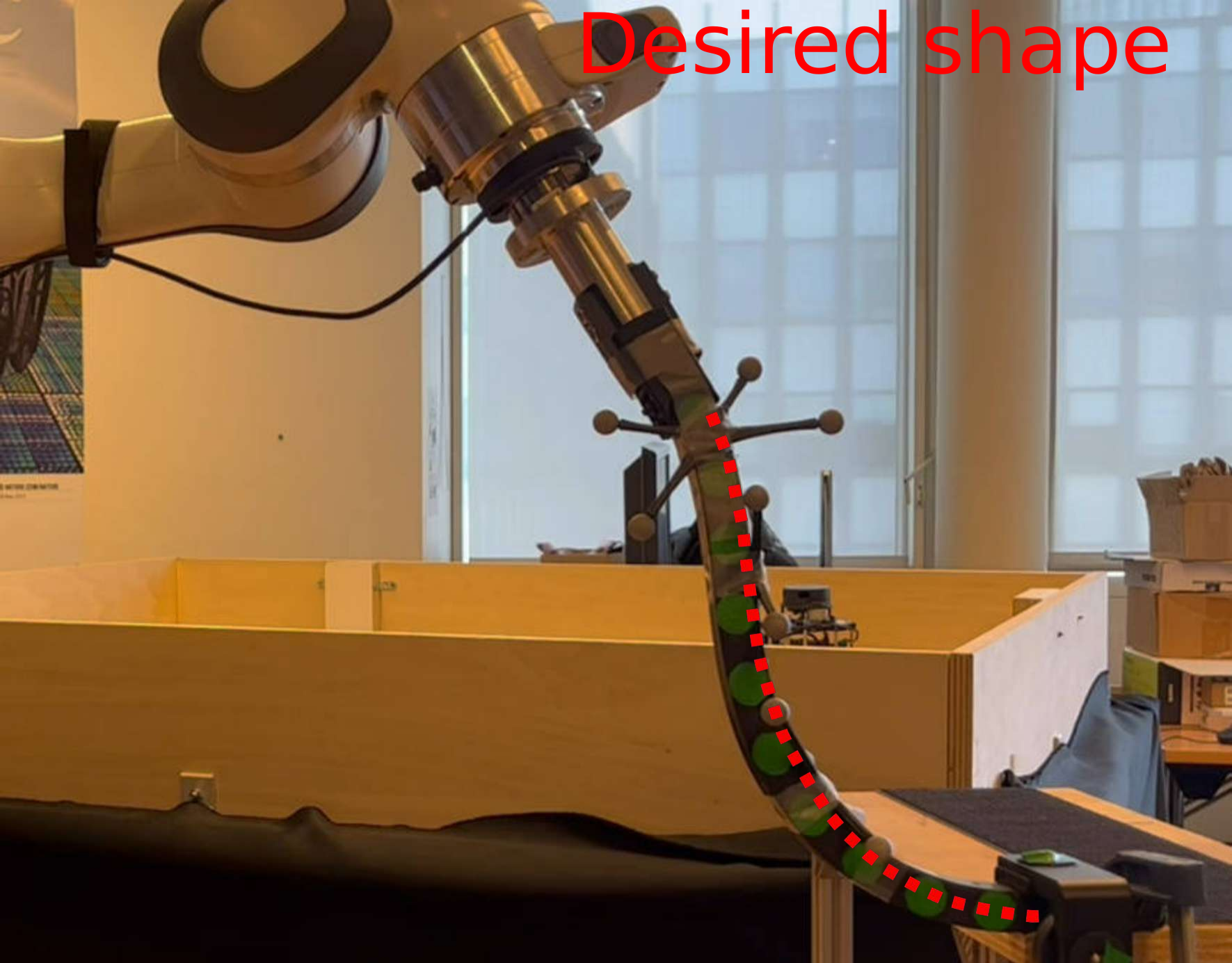}}
    \subfloat[Desired shape]
    {\includegraphics[width=0.22\textwidth]{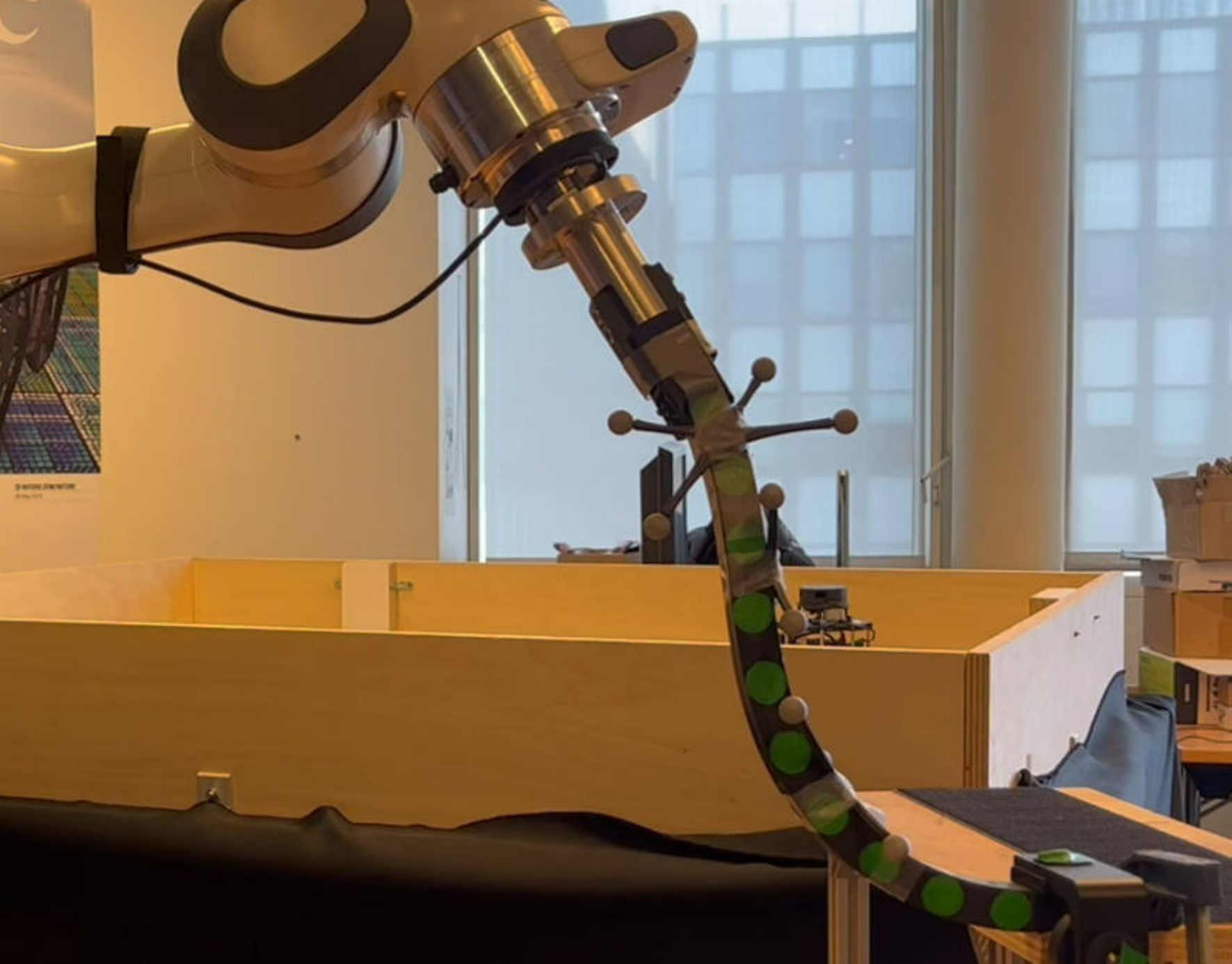}}
    \subfloat[Error over time]
    {\includegraphics[width=0.22\textwidth]{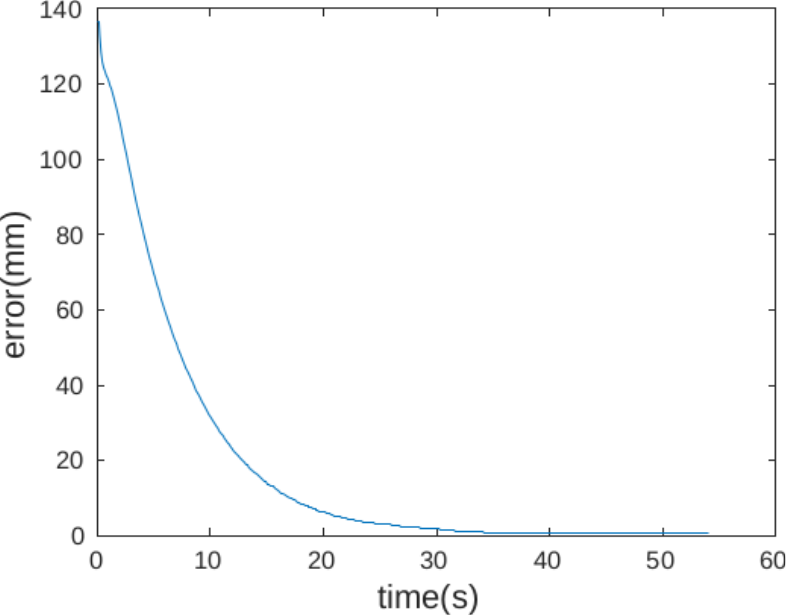}}
    
    \caption{Experiment with different boundary conditions: The object is grasped at one limit and clamped at the other}
    \label{fig: one arm}
\end{figure*}

We followed the same protocol as proposed in experiment 1, however in this case we only had one arm manipulating the object, which is clamped at the other limit. 
This changes the boundary conditions of the problem, thus adding additional constraints on the position and orientation of the object at $s=0$. The disturbance $\Delta^{j}_{d}$ introduced along these dimensions in $\Gamma_{0}$ will therefore be equal to zero when computing the Jacobian. 
To avoid computing the pseudo-inverse of sparse matrices, we reduce the dimension of the problem to the sole variation of $n_{0}$ and $m_{0}$ in $\Gamma_{0}$. This effectively reduces the number of variables by half. The comparison between the initial, final, and desired shapes is represented in figure \ref{fig: one arm}. 

\subsection{Comparison with the state of the art and Discussion}

 

\begin{table*}[hbt]
\centering
\begin{tabular}{|p{2.1cm}||p{1.3cm}|p{1.4cm}|p{1.6cm}|p{1.5cm}|p{1.6cm}|p{2.8cm}|p{1.5cm}|}
 \hline
 Method & Physically realistic & Precision & Convergence & Loop frequency & Shape complexity & Material & Object type\\
 \hline
 
O-L Cosserat \cite{azad2023optimal}       &Yes& $5$ mm   & Linear       & Open-loop & 3 & Isotropic+homogeneous & Linear\\
ARAP-SS \cite{shetab2022lattice}    &No & $5$ mm   & Linear       & 20 Hz & 3 & Isotropic+homogeneous & Linear+Planar\\      
DLO NN \cite{caporali2024deformable}&No & $10$ mm     & Linear       & 1500 Hz & 3 & Isotropic+homogeneous& Linear\\
FEM \cite{koessler2021efficient}    & Yes & $20$ mm   & Exponential  & 30 Hz   & 2 & Isotropic+homogeneous& Linear\\
Our approach                                    &Yes& $<1$ mm   & Exponential  & $>10$ Hz & 3 & Any& Linear\\
 \hline
\end{tabular}
\caption{Comparison of the shape servoing of deformable linear objects (DLO) methods}
\label{table comparison}
\end{table*}

A comparison of the most promising approaches including open-loop Cosserat for robotic shape control of linear deformable objects is displayed on table \ref{table comparison}. We evaluate several factors: from the physical realism of the obtained shape (which ensures feasibility) to the shape complexity corresponding to the highest order of the Legendre polynomial that approximates the proposed shapes. The only area where our algorithm falls behind is the loop frequency which can be easily sped up by changing the optitrack system with another vision-based approach (for reference the loop frequency without the vision is 100 Hz). On the other side, our approach is the only one that consistently reaches sub-millimeter precision for such a range of complex desired shapes, even when manipulating objects with complex compositions and unknown properties.

The closest method in term of performance is the ARAP-SS algorithm proposed in \cite{shetab2022lattice}. To further enhance the comparison, we introduce the same normalized metric based on cosine similarity between the measured positions of the control points and their predicted positions. This metric represents the alignment between the predicted and measured vectors. The nearer the value is to 1, the more accurately the deformation Jacobian reflects the behavior of the manipulated object. The mean between the cosine similarities of each tests as well as the inter-test standard deviation is exposed on table \ref{table comparison cosine}.

\begin{table}[hbt]
\centering
\begin{tabular}{|p{0.9cm}||p{1.6cm}|p{1.6cm}|p{1.6cm}|p{0.8cm}|}
 \hline
Method & Our (object 1)  & Our (object 2) & Our (object 3) & ARAP\\
 \hline
mean & 0.9884 & 0.9792 & 0.9771 & 0.8483 \\
std  & 0.0082 & 0.0349 & 0.0154 & 0.0379 \\
 \hline
\end{tabular}
\caption{Average cosine similarity and standard deviation over all tests for each object and method}
\label{table comparison cosine}
\end{table}

\noindent Our approach shows higher mean cosine similarity for all objects than the ARAP-SS method. The standard deviation can directly be tied to the robustness of the approach as it shows a regularity of the metric across different tests. 
The difference in mean and standard deviation between the first object and the others reflects the imprecision on the stiffness parameters. 
Nevertheless, we can conclude that the Jacobian accurately represents the behavior of the manipulated object in all scenarios. The Jacobian primarily captures the direction of deformations, which necessitates precise modeling. On the other side, the stiffness parameters mainly influence the amplitude of said deformations, which can be readily adjusted through closed-loop control. This explains the robustness of the approach to stiffness estimation.

However, the method has limitations. Firstly, although it is general amond DLOs, the Cosserat rod model used in this article is specifically designed for one-dimensional objects and is therefore not suitable for modeling 2D or 3D shapes. Secondly, the method does not accommodate unforeseen contacts that have not been predefined in the model. Managing unexpected contacts remains an open challenge in the field of deformable object manipulation.




\section{Conclusion}

\noindent We have presented in this article our method to control the shape of DLOs in real-time. By defining our global BVP and iteratively solving an IVP, we included the Cosserat rod model in our closed-loop controller. Thus enhancing the precision compared to regular open-loop physics-based approaches while also demonstrating robustness to stiffness estimation. 
We then conducted three different experiments, showcasing the precision and repeatability of the approach for various desired shapes. The approach consistently achieved sub-millimeter precision during shape control tests on various objects with anisotropic and heterogeneous properties.
An interesting perspective would be to replace the perception module with more advanced methods to detect and estimate contacts accurately. The contact forces could then be included in the model to allow the approach to handle unexpected contacts. Finally, the approach can be generalized to other applications such as deformable linear robots control. The problem's formulation varies depending on the type of actuation used. However, it is generally simpler because the position and orientation are fixed at the robot's base. Consequently, the main challenge is accurately defining the relationship between the variation of internal forces and moments at $s=0$ and the actuation.


\bibliographystyle{IEEEtran}
\bibliography{biblio}

\end{document}